\documentclass[lettersize,journal]{IEEEtran}
\usepackage{amsmath,amsfonts}
\usepackage{algorithmic}
\usepackage{algorithm}
\usepackage{array}
\usepackage[caption=false,font=normalsize,labelfont=sf,textfont=sf]{subfig}
\usepackage{textcomp}
\usepackage{stfloats}
\usepackage{xcolor}
\usepackage{adjustbox}
\usepackage{tikz}
\usepackage{hyperref}
\hypersetup{
    colorlinks=true,
    allcolors=blue
}
\newcommand{\bluecite}[1]{{\color{blue}[\citenum{#1}]}}
\usepackage{url}
\usepackage{bm}
\usepackage[edges]{forest}
\usepackage{booktabs}
\usepackage{multirow}
\usepackage{multicol}
\usepackage{makecell}
\usepackage{verbatim}
\usepackage{graphicx}
\usepackage{colortbl} 
\usepackage{cite}
\usepackage{pifont}
\usepackage{fontawesome5}
\usepackage{cleveref}
\usepackage{enumitem}
\definecolor{mattered}{RGB}{214, 26, 60}
\definecolor{mattegreen}{HTML}{369F39}

\usepackage{bbding}

\definecolor{hidden-draw}{RGB}{20,68,106}
\definecolor{hidden-pink}{RGB}{255,245,247}
\definecolor{my_green}{RGB}{51,102,0}
\definecolor{my_red}{RGB}{204, 0, 0}
\definecolor{st_color}{RGB}{0, 90, 180}
\definecolor{t_color}{RGB}{255, 102, 102}

\newcommand{\cmark}{\textcolor{my_green}{\ding{51}}} 
\newcommand{\xmark}{\textcolor{my_red}{\ding{55}}} 

\DeclareMathOperator*{\argmax}{arg\,max}

\begin{document}

\title{Video Understanding with Large Language Models:\\ A Survey}

\author{
Yunlong Tang, Jing Bi, Siting Xu, Luchuan Song, Susan Liang,~\IEEEmembership{Graduate Student Member,~IEEE}, Teng Wang, Daoan Zhang, Jie An, Jingyang Lin, Rongyi Zhu, Ali Vosoughi,~\IEEEmembership{Graduate Student Member,~IEEE}, Chao Huang, Zeliang Zhang, Pinxin Liu, Mingqian Feng, Feng Zheng,~\IEEEmembership{Member,~IEEE}, Jianguo Zhang,~\IEEEmembership{Senior Member,~IEEE}, Ping Luo,~\IEEEmembership{Member,~IEEE}, Jiebo Luo,~\IEEEmembership{Fellow,~IEEE}, and Chenliang Xu,~\IEEEmembership{Member,~IEEE}
\thanks{Y. Tang, J. Bi, L. Song, S. Liang, D. Zhang, J. An, J. Lin, R. Zhu, A. Vosoughi, C. Huang, Z. Zhang, P. Liu, M. Feng, J. Luo, and C. Xu are with University of Rochester}
\thanks{T. Wang and P. Luo are with The University of Hong Kong}
\thanks{S. Xu, T. Wang, F. Zheng, and J. Zhang are with Southern University of Science and Technology}
\thanks{Corresponding to Y. Tang, J. Luo, and C. Xu (\{yunlong.tang@, jluo@cs., chenliang.xu@\}rochester.edu)}
}



\maketitle

\begin{abstract}
With the rapid growth of online video platforms and the escalating volume of video content, the need for proficient video understanding tools has increased significantly. 
Given the remarkable capabilities of large language models (LLMs) in language and multimodal tasks, this survey provides a detailed overview of recent advances in video understanding that harness the power of LLMs (Vid-LLMs).
The emergent capabilities of Vid-LLMs are surprisingly advanced, particularly their ability for open-ended multi-granularity (abstract, temporal, and spatiotemporal) reasoning combined with common-sense knowledge, suggesting a promising path for future video understanding.
We examine the unique characteristics and capabilities of Vid-LLMs, categorizing the approaches into three main types: \textit{Video Analyzer $\times$ LLM}, \textit{Video Embedder $\times$ LLM}, and \textit{(Analyzer + Embedder) $\times$ LLM}. We identify five subtypes based on the functions of LLMs in Vid-LLMs: \textit{LLM as Summarizer}, \textit{LLM as Manager}, \textit{LLM as Text Decoder}, \textit{LLM as Regressor}, and \textit{LLM as Hidden Layer}.
This survey also presents a comprehensive study of the tasks, datasets, benchmarks, and evaluation methods for Vid-LLMs.
Additionally, it explores the extensive applications of Vid-LLMs in various domains, highlighting their remarkable scalability and versatility in real-world video understanding challenges.
Additionally, it summarizes the limitations of existing Vid-LLMs and outlines directions for future research. For more information, readers are encouraged to visit the repository at \href{https://github.com/yunlong10/Awesome-LLMs-for-Video-Understanding}{https://github.com/yunlong10/Awesome-LLMs-for-Video-Understanding}.

\end{abstract}


\section{Introduction}

\begin{figure*}[!h]
    \centering
    \includegraphics[width=0.95\linewidth]{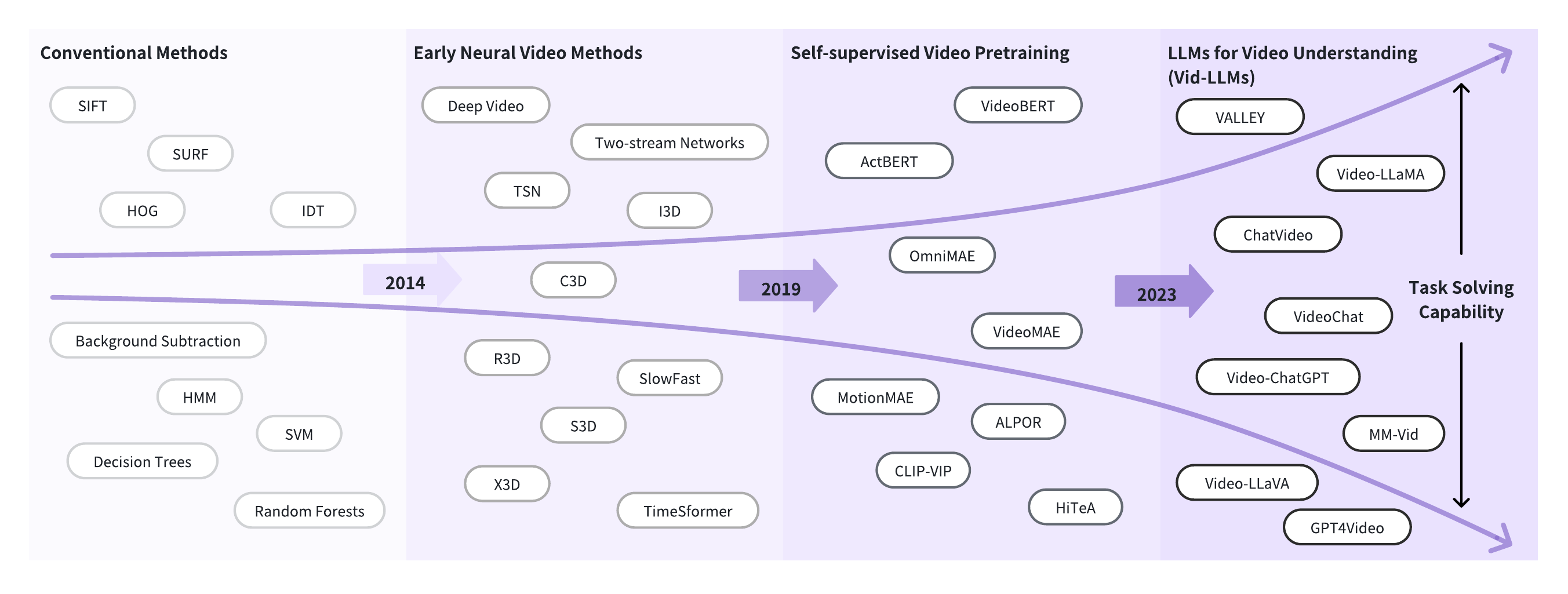}
    \caption{The development of video understanding methods can be summarized into four stages: (1) Conventional Methods, (2) Early Neural Video Models, (3) Self-supervised Video Pretraining, and (4) Large Language Models for Video Understanding, i.e., Vid-LLMs. Their task-solving capability is continuously improving, and they possess the potential for further enhancement.}
    \label{fig:milestone}
\vspace{-1em}
\end{figure*}

\IEEEPARstart{W}e live in a multimodal world where video has become the predominant form of media. With the rapid expansion of online video platforms and the growing prevalence of cameras in surveillance, entertainment, and autonomous driving, video content has risen to prominence as a highly engaging and rich medium, outshining traditional text and image-text combinations in both depth and appeal.
This advancement has fueled an exponential increase in video production, with millions of videos being created every day.
However, manually processing such a sheer volume of video content is labor-intensive and time-consuming.
As a result, there is a growing need for tools to effectively manage, analyze, and process this abundance of video content.
To meet this need, video understanding methods have emerged that use intelligent analysis techniques to automatically recognize and interpret video content, significantly reducing the workload on human operators.
In addition, the ongoing development of these methods is improving their task-solving capabilities, enabling them to handle a wide range of video understanding tasks with increasing proficiency.

\subsection{Development of Video Understanding Methods}
The evolution of video understanding methods can be divided into four stages, as shown in Figure~\ref{fig:milestone}:

\subsubsection{Conventional Methods}
In the early stages of video understanding, handcrafted feature extraction techniques such as Scale-Invariant Feature Transform (SIFT)~\bluecite{lindeberg2012scale}, Speeded-Up Robust Features (SURF)~\bluecite{bay2008speeded}, and Histogram of Oriented Gradients (HOG)~\bluecite{dalal2005histograms} were used to capture key information in videos.
Background Subtraction~\bluecite{sobral2014comprehensive}, optical flow methods~\bluecite{tu2022optical}, and Improved Dense Trajectories (IDT)~\bluecite{wang2013action, shu2015action} were used to model the motion information for tracking. 
Since videos can be viewed as time series data, temporal analysis techniques such as Hidden Markov Models (HMM)~\bluecite{liu2003video} have also been used to understand video content.
Before the popularity of deep learning, basic machine learning algorithms such as Support Vector Machines (SVM)~\bluecite{sidenbladh2004detecting}, Decision Trees~\bluecite{yuan2002automatic}, and Random Forests were also used in video classification and recognition tasks.
Cluster analysis~\bluecite{chan2008modeling} for classifying video segments, or Principal Component Analysis (PCA)~\bluecite{bouwmans2014robust, hazelhoff2008video} for data dimensionality reduction have also been commonly used methods for video analysis.
    
\subsubsection{Early Neural Video Models}
Compared with classical methods, deep learning methods for video understanding possess superior task-solving capabilities. DeepVideo~\bluecite{KarpathyCVPR14} and \bluecite{earliest} were early methods introducing a deep neural network, specifically a Convolutional Neural Network (CNN), for video understanding. However, the performance was not superior to the best handcrafted feature method due to the inadequate use of motion information. Two-stream networks~\bluecite{feichtenhofer2016convolutional} combined both CNN and IDT to capture the motion information to improve the performance, which verified the capability of deep neural networks for video understanding. To handle long-form video understanding, Long Short-Term Memory (LSTM) was adopted~\bluecite{yue2015beyond}. Temporal Segment Network (TSN)~\bluecite{wang2016temporal} was also designed for long-form video understanding by analyzing and aggregating video segments. Besides TSN, Fisher Vectors (FV) encoding~\bluecite{sekma2015human}, Bi-Linear encoding~\bluecite{diba2017deep}, and Vector of Locally Aggregated Descriptors (VLAD)~\bluecite{mironicua2016modified} encoding were introduced~\bluecite{li2022transvlad}. These methods improved performance on the UCF-101~\bluecite{soomro2012ucf101} and HMDB51~\bluecite{kuehne2011hmdb} datasets. Unlike two-stream networks, 3D networks started another branch by introducing 3D CNN to video understanding (C3D)~\bluecite{tran2015learning}. Inflated 3D ConvNets (I3D)~\bluecite{carreira2017quo} utilizes the initialization and the architecture of 2D CNN, Inception~\bluecite{szegedy2015going}, to gain a huge improvement on the UCF-101 and HMDB51 datasets. Subsequently, people began employing the Kinetics-400 (K-400)~\bluecite{kay2017kinetics} and Something-Something~\bluecite{goyal2017something} datasets to evaluate the model's performance in more challenging scenarios. ResNet~\bluecite{he2016deep}, ResNeXt~\bluecite{Xie_2017_CVPR}, and SENet~\bluecite{Hu_2018_CVPR} were also adapted from 2D to 3D, resulting in the emergence of  R3D~\bluecite{Hara_2017_ICCV}, MFNet~\bluecite{Chen_2018_ECCV}, and STC~\bluecite{Diba_2018_ECCV}. To improve the efficiency, the 3D convolution has been decomposed into cascade 2D and 1D convolution in various studies (e.g., S3D~\bluecite{S3D}, ECO~\bluecite{zolfaghari2018eco}, P3D~\bluecite{qiu2017learning}). LTC~\bluecite{LTC}, T3D~\bluecite{T3D}, Non-local~\bluecite{non-local}, and V4D~\bluecite{zhang2019v4d} focus on long-form temporal modeling, while CSN~\bluecite{tran2019video}, SlowFast~\bluecite{feichtenhofer2019slowfast}, and X3D~\bluecite{feichtenhofer2020x3d} tend to attain high efficiency. The introduction of Vision Transformers (ViT)~\bluecite{dosovitskiy2021an} promotes a series of prominent models (e.g., TimeSformer~\bluecite{bertasius2021space}, VidTr~\bluecite{li2021vidtr}, ViViT~\bluecite{arnab2021vivit}, MViT~\bluecite{fan2021multiscale}).

\begin{figure*}[!h]
    \centering
    \includegraphics[width=0.95\textwidth]{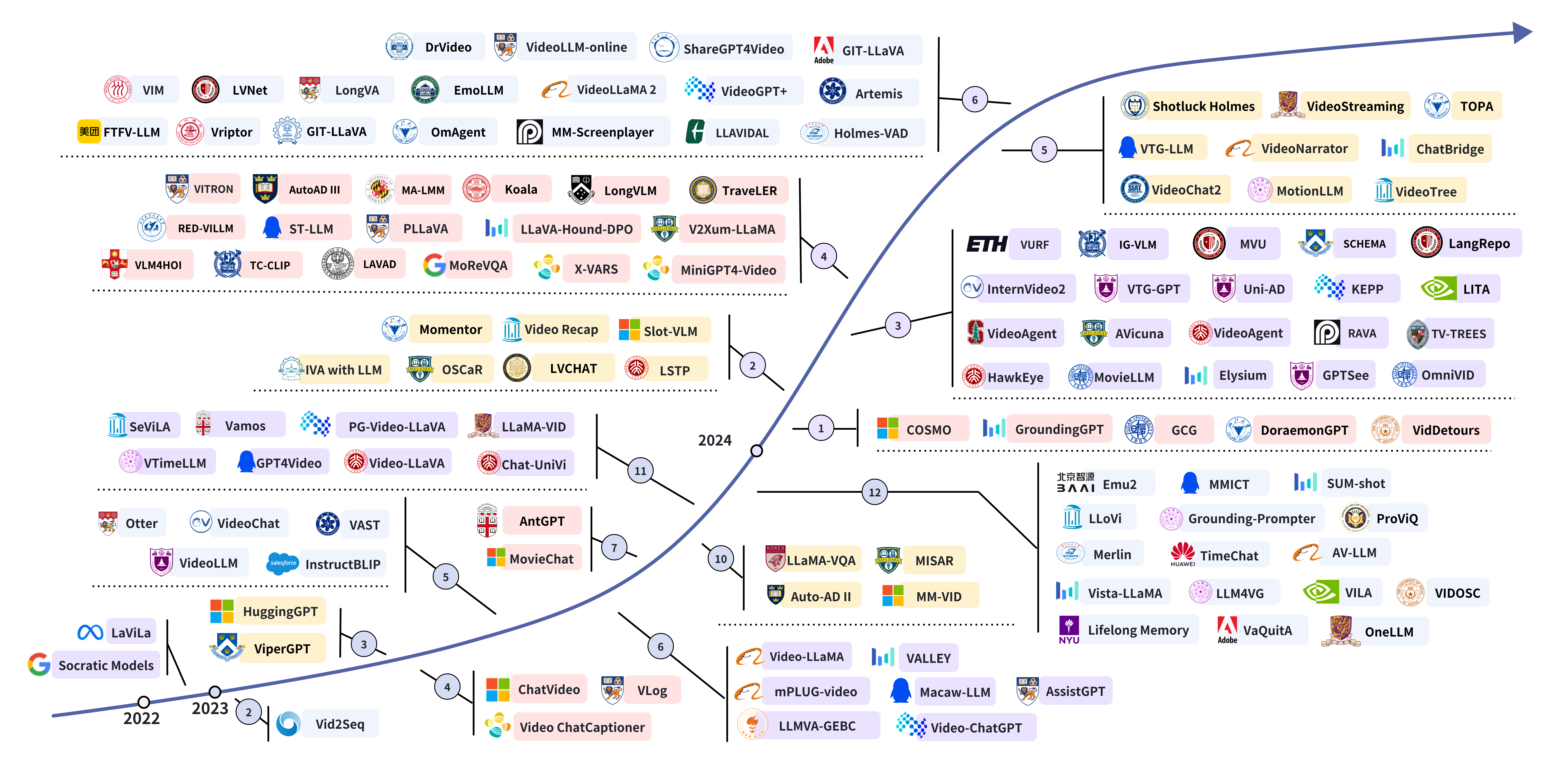}
    \caption{A comprehensive timeline depicting the development of video understanding methods with large language models~(Vid-LLMs). This survey is based on advancements up to the end of June 2024.}
    \label{fig:timeline}
\vspace{-1em}
\end{figure*}

\subsubsection{Self-supervised Video Pretraining}
Transferability~\bluecite{li2023cross, wang2023feature} in self-supervised pretraining models~\bluecite{zhang2023rethinking} for video understanding allows them to generalize across diverse tasks with minimal additional labeling, overcoming the early deep learning models' requirements for extensive task-specific data.
    VideoBERT~\bluecite{sun2019videobert} is an early attempt to perform video pretraining. Based on the bidirectional language model BERT~\bluecite{kenton2019bert}, pertaining tasks are designed for self-supervised learning from video-text data. It tokenizes video features with hierarchical k-means. The pretrained model can be fine-tuned to handle multiple downstream tasks, including action classification and video captioning. Following the \textit{``pretraining-finetuning"} paradigm, many studies on pretrained models for video understanding, especially video-language models, have emerged. They either use different architectures (ActBERT~\bluecite{zhu2020actbert}, SpatiotemporalMAE~\bluecite{feichtenhofer2022masked}, OmniMAE~\bluecite{girdhar2023omnimae}, VideoMAE~\bluecite{tong2022videomae}, MotionMAE~\bluecite{yang2022self}) or training strategies (MaskFeat~\bluecite{wei2022masked}, VLM~\bluecite{xu2021vlm}, ALPRO~\bluecite{li2022align}, All-in-One transformer~\bluecite{moritz2020all}, MaskViT~\bluecite{gupta2022maskvit}, CLIP-ViP~\bluecite{xue2022clip}, Singularity~\bluecite{lei2022revealing}, LF-VILA~\bluecite{sun2022long}, EMCL~\bluecite{jin2022expectation}, HiTeA~\bluecite{ye2023hitea}, CHAMPAGNE~\bluecite{han2023champagne}).

\subsubsection{Large Language Models for Video Understanding}
Recently, large language models (LLMs) have advanced rapidly \bluecite{lyu2023gpt}. The emergence of large language models pretrained on extensive datasets has introduced a novel in-context learning capability \bluecite{zhang2023dnagpt}. This allows them to handle various tasks using prompts without the need for fine-tuning. ChatGPT~\bluecite{OpenAIChatGPT} is the first groundbreaking application built on this foundation. This includes capabilities like generating code and invoking tools or APIs of other models for their use. Many studies are exploring using LLMs like ChatGPT to call vision models APIs to solve the problems in the computer vision field, including Visual-ChatGPT~\bluecite{wu2023visual}. The advent of instruct-tuning has further enhanced these models' ability to respond effectively to user requests and perform specific tasks~\bluecite{liu2023visual}. LLMs integrated with video understanding capabilities offer the advantage of more sophisticated multimodal understanding, enabling them to process and interpret complex interactions between visual and textual data. Similar to their impact in Natural Language Processing (NLP)~\bluecite{zhao2023survey}, these models act as more general-purpose task solvers, adept at handling a broader range of tasks by leveraging their extensive knowledge base and contextual understanding acquired from vast amounts of multimodal data. This allows them to not only understand visual content but also reason about it in a way that is more aligned with human-like understanding. Many works also explore using LLMs in video understanding tasks, namely, Vid-LLMs.


\subsection{Related Surveys}
Previous survey papers either study specific sub-tasks in the area of video understanding or focus on methodologies beyond video understanding. For example, \bluecite{li2023multimodal} surveys multimodal foundation models for general vision-language tasks, which includes both image and video applications. \bluecite{survey_vid_cap} and \bluecite{survey_vid_act_reg} focus on surveying video captioning and action recognition tasks, respectively. Other video understanding tasks, such as the video question answering and grounding, are not considered. Moreover, \bluecite{survey_vdm}, \bluecite{annepaka2024large}, and \bluecite{zhao2023survey} survey video-related methodologies, such as video diffusion models and LLMs, lacking the concentration on video understanding.
\bluecite{madan2024foundation} centers primarily on video foundation models, with insufficient attention given to language model-based approaches.
Despite the significant value to the community, previous survey papers leave a gap in surveying the general video understanding task based on large language models.
This paper fills this gap by comprehensively surveying the video understanding task using large language models. 


\subsection{Survey Structure}
This survey is structured as follows: 
Section~\ref{sec:preliminary} offers preliminaries for video understanding with LLMs, including a summary of various video understanding tasks that require handling different levels of granularity, their associated datasets, and evaluation metrics. The background of LLMs is also introduced in this section.
In Section \ref{sec:model}, we delve into details of recent research leveraging LLMs for video understanding, presenting their unique approaches and impact in the field, where we divide these Vid-LLMs into three main categories, \textit{Video Analyzer $\times$ LLM} and \textit{Video Embedder $\times$ LLM}, and \textit{(Analyzer + Embedder) $\times$ LLM}; and five sub-categories, \textit{LLM as Summarizer/Manager/Text Decoder/Regressor/Hidden Layer}, shown as \Cref{fig:taxonomy}. This section also includes the training strategies of Vid-LLMs.
Section~\ref{sec:benchmark_eval} adds more information about popular ways to evaluate Vid-LLMs, together with some benchmarks and performances of some Vid-LLMs on the most commonly used benchmarks.
Section~\ref{sec:sum} explores the application of Vid-LLMs across multiple significant fields and identifies unresolved challenges and potential areas for future research.

In addition to this survey, we have established a GitHub repository that aggregates various supporting resources for video understanding with large language models (Vid-LLMs). This repository, dedicated to enhancing video understanding through Vid-LLMs, can be accessed at \href{https://github.com/yunlong10/Awesome-LLMs-for-Video-Understanding}{\textit{Awesome-LLMs-for-Video-Understanding}}.

\section{Preliminaries}
\label{sec:preliminary}
In this section, we introduce the background of video understanding and Large Language Models (LLMs).
\subsection{Video Understanding Tasks}
Video understanding is a fundamental yet challenging task that has inspired the emergence of numerous tasks in a similar discipline aiming at interpreting complicated video content. 
The pioneering work for video understanding includes video classification and action recognition approaches, which classify videos into class labels and action categories, respectively. 
With the development of visual foundation models and expanding public datasets, current video understanding approaches can capture, analyze, and reason for more complicated video content. 
For instance, video captioning, as a specific task of video understanding, not only requires the model to generate detailed descriptions of the video content, but the generated video captions should be logical and follow commonsense about the scenes depicted. 
Additionally, the Video Question-Answering (VQA) task requires that the model understand the content and refer to external information to provide an accurate answer. 
The development path of video understanding from simple classification to natural language comprehension and reasoning highlights a clear trend of the video understanding model towards near-human levels of video interpretation capability. We summarize the main tasks in video understanding as follows.

\begin{figure*}
    \centering
    \includegraphics[width=0.95\linewidth]{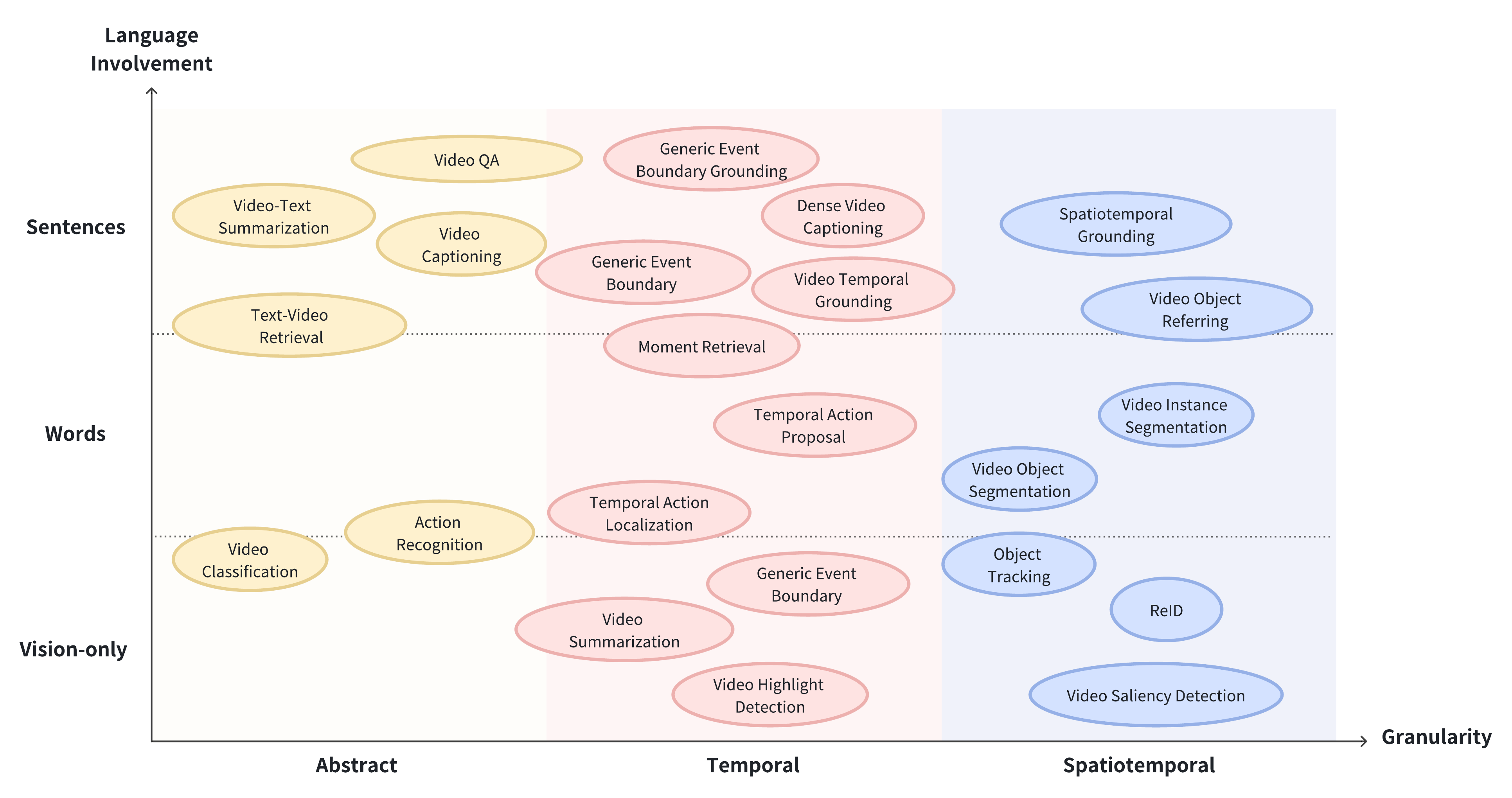}
    \caption{This figure categorizes tasks in video understanding, delineating the granularity required and the language involvement necessary for models to perform these tasks effectively. This diagram excludes tasks involving special modalities or specific, such as audio-visual and egocentric video understanding. Notably, the tasks presented could be unified into a question-answering paradigm, and all solved by generative large models, akin to recent advances in NLP.}
    \label{fig:tasks}
\vspace{-1em}
\end{figure*}

\subsubsection{Abstract Understanding Tasks}
Video Classification, Action Recognition, Text-Video Retrieval, Video-to-Text Summarization, and Video Captioning.

\begin{itemize}
    \item \textit{Video Classification \& Action Recognition:} Video classification and action recognition classify videos based on class labels or activities and events categories within a video sequence. Datasets specifically introduced for these tasks include UCF-101~\bluecite{soomro2012ucf101}, HMDB51~\bluecite{kuehne2011hmdb}, Hollywood~\bluecite{hollywood2}, {ActivityNet}~\bluecite{caba2015activitynet}, {Charades}~\bluecite{sigurdsson2016hollywood}, {Kinetics-400}~\bluecite{kay2017kinetics}, {Kinetics-600}~\bluecite{carreira2018short}, Kinetics-700~\bluecite{carreira2019short}, SomethingSomethingV2~\bluecite{goyal2017something}, HACS~\bluecite{zhao2019hacs}, {YouTube8M}~\bluecite{abuelhaija2016youtube8m}, and PortraitMode-400~\bluecite{han2024video_portraitmode-400}. Usually, Top-K accuracy is adopted as the main metric for these tasks.
    
    \item \textit{Text-Video Retrieval:} Text-video retrieval task matches and retrieves relevant video clips based on the similarity between video clips and the input textual descriptions. Datasets like Kinetic-GEB~\bluecite{wang2022geb+}, MSRVTT~\bluecite{xu2017video}, DiDeMo~\bluecite{anne2017localizing}, YouCook2~\bluecite{zhou2018towards}, and {Youku-mPLUG}~\bluecite{xu2023youku} are relevant to this task. The standard evaluation metric for this task is Recall at K (R@K), which measures the accuracy of the first K retrieved results.
    
    \item \textit{Video-to-Text Summarization:} Video-to-text summarization is a task that generates concise textual summaries of videos. The video summarization approaches are trained to extract and interpret key visual and audio content to produce coherent and informative summaries. 
    ViTT~\bluecite{huang2020multimodal}, {VideoXum} \bluecite{lin2023videoxum}, {VideoInstruct-100K}~\bluecite{maaz2023video}, and {Instrcut-V2Xum}~\bluecite{hua2024v2xum} are datasets that related to the task. Metrics of BLEU, METEOR, CIDEr, and ROUGE-L often evaluate this task. 
    
    \item \textit{Video Captioning:} Video captioning generates descriptive and coherent textual captions of given videos. The video caption models usually use visual and auditory information from video to produce accurate and contextually relevant descriptions. 
    Notable datasets for this tasks are MSVD~\bluecite{chen2011collecting}, MSR-VTT~\bluecite{xu2016msr}, TGIF~\bluecite{li2016tgif}, {Charades}~\bluecite{sigurdsson2016hollywood}, {Charades-Ego}~\bluecite{sigurdsson2018charades}, {YouCook2}~\bluecite{zhou2018towards}, {Youku-mPLUG}~\bluecite{xu2023youku}, {VAST-27M}~\bluecite{chen2023vast}, and {VideoInstruct-100K}~\bluecite{maaz2023video}. This task is often evaluated by metrics of BLEU, METEOR, CIDEr, and ROUGE-L. 
    
    \item \textit{Video QA:} Video Question-Answering (VQA) aims to answer textual questions based on a given video, where the model analyzes visual and auditory information, understands the context, and eventually generates accurate responses.
    Datasets involved in the QA task are VCR~\bluecite{zellers2019recognition}, {MSVD-QA} \bluecite{xu2017video}, {MSRVTT-QA} \bluecite{xu2017video}, {TGIF-QA} \bluecite{jang2017tgif},  {Pororo-QA}~\bluecite{kim2017deepstory}, {TVQA} \bluecite{lei2018tvqa}, {ActivityNet-QA}~\bluecite{yu2019activitynet}, and NExT-QA~\bluecite{NExT-QA}. This task is evaluated using Top-1, Top-K accuracy.
\end{itemize}

\subsubsection{Temporal Understanding Tasks}

\begin{itemize}
    \item \textit{Video Summarization:} Video Summarization aims at condensing a long video into a shorter version while preserving essential content. F1-score, Spearman, and Kendall usually evaluate this task as metrics. Commonly-used datasets include {SumMe}~\bluecite{gygli2014creating}, {TVSum}~\bluecite{song2015tvsum}, Ads-1k~\bluecite{tang2022multi}, {VideoXum}~\bluecite{lin2023videoxum}, {Instrcut-V2Xum}~\bluecite{hua2024v2xum}. 
    
    \item \textit{Video Highlight Detection:} Video highlight detection aims at identifying and extracting the most important and interesting segments from a video. Commonly used datasets on this task include the YouTube Highlights~\bluecite{youtubeHighlights}, the TV-Sum~\bluecite{song2015tvsum}, and the Videos Titles in the Wild (VTW)~\bluecite{VTWdataset}.
    
    \item \textit{Temporal Action/Event Localization:} This task aims at identifying the precise temporal segments of actions or events within a video. By analyzing sequential frames, models trained for this task must indicate when specific activities start and end. Datasets for Temporal Action/Event Localization involve THUMOS'14~\bluecite{THUMOS14}, {ActivityNet-1.3}~\bluecite{caba2015activitynet}, and UnAV-100~\bluecite{geng2023dense_unav100}.
    
    \item \textit{Temporal Action Proposal Generation:} Temporal action proposal generation involves generating candidate segments within a video that are likely to contain actions or events. Relevant datasets like THUMOS'14~\bluecite{THUMOS14}, {ActivityNet}~\bluecite{caba2015activitynet}, and {Charades}~\bluecite{sigurdsson2016hollywood} are used for the training and evaluation for this task.
    
    \item \textit{Video Temporal Grounding:} Video temporal grounding is the task of locating specific moments or intervals within a video that correspond to a given textual query. This process involves aligning linguistic descriptions with visual content, enabling precise identification of relevant segments for applications in video search and content analysis. 
    Common benchmarks are {Charades-STA}~\bluecite{Gao_2017_ICCV}, ViTT~\bluecite{huang2020multimodal}, {DiDeMo}~\bluecite{anne2017localizing}, and PU-VALOR~\bluecite{tang2024avicuna}. The metrics of R1@0.5 and R1@0.7 often evaluate this task.
 
    \item \textit{Moment Retrieval:} Moment retrieval is the task of identifying and extracting precise video segments that correspond to a given textual or visual query, which aligns semantic content between queries and video frames. {DiDeMo}~\bluecite{anne2017localizing} is a dataset for this task. 

    \item \textit{Generic Event Boundary Detection:} Generic event boundary detection involves identifying certain frames in a video where significant changes occur and splitting videos based on different events or activities. Kinetics-GEBD~\bluecite{shou2021generic} is a widely-used dataset for this task. 
    
    \item \textit{Generic Event Boundary Captioning \& Grounding:} Generic event boundary captioning and grounding involve identifying and describing the transition points between significant events in a video, where Kinetics-GEB+~\bluecite{wang2022geb+} is the dataset for this task.  
    
    \item \textit{Dense Video Captioning:} Dense video captioning~\bluecite{shao2022region,Wang_2021_ICCV,shao2023textual,long2023capdet,shao2024dcmstrd} aims at generating detailed and continuous textual descriptions for multiple events and actions occurring throughout a video. Evaluation metrics like BLEU, METEOR, CIDEr, and ROUGE-L are used to evaluate this task. Relevant datasets are {ActivityNet Captions}~\bluecite{krishna2017dense}, {VidChapters-7M}~\bluecite{yang2023vidchapters}, {YouCook2}~\bluecite{zhou2018towards}, and ViTT~\bluecite{huang2020multimodal}. 
\end{itemize}

\subsubsection{Spatiotemporal Understanding Tasks}
\begin{itemize}
    \item \textit{Object Tracking:} Object tracking aims at continuously identifying and following the position of specific objects within a video over time. A good tracking model should maintain accurate and consistent trajectories of objects, even for videos with occlusions, changes in appearance, and motions. Benchmarks like OTB~\bluecite{OTBbenchmark}, UAV~\bluecite{UAVbenchmark}, and VOT~\bluecite{VOTbenchmark} are commonly used for this task. 
    
    \item \textit{Re-Identification:} Re-Identification (ReID) is the task of recognizing and matching individuals or objects across different video frames or camera views. Common datasets in ReID are Market-1501~\bluecite{Market1501}, CUHK03~\bluecite{cuhk03}, MSMT17~\bluecite{wei2018person}, and DukeMTMC-reID~\bluecite{dukeMTMCreID}.
    
    \item  \textit{Video Saliency Detection:} Video saliency detection aims at identifying the most visually important and attention-grabbing regions in a video~\bluecite{moskalenko2024aim}. This task highlights areas that stand out due to factors like motion, contrast, and unique features. 
    Relevant datasets to this task are DHF1K~\bluecite{wang2018revisiting_dhf1k}, Hollywood-\bluecite{sigurdsson2016hollywood}, UCF-Sports~\bluecite{ucfsport}, AVAD~\bluecite{avad_sal}, Coutrot1~\bluecite{coutrot1_2014saliency}, Coutrot2~\bluecite{coutrot2_2016multimodal}, ETMD~\bluecite{ETMD_koutras2015perceptually}, and {SumMe}~\bluecite{gygli2014creating}.
    
    \item \textit{Video Object Segmentation:} Video object segmentation aims at partitioning a video into segments that correspond to individual objects, accurately delineating their boundaries over time. YouTube-VOS~\bluecite{youtubevos} and DAVIS~\bluecite{perazzi2016benchmark} are datasets related to this task. 
    
    \item \textit{Video Instance Segmentation:} Video instance segmentation is the task of identifying, segmenting, and tracking each unique object instance within a video. YouTube-VIS~\bluecite{youtubevis} and Cityscapes-Seq~\bluecite{cityscape} are two common benchmarks for this task. 
    
    \item \textit{Video Object Referring Segmentation:} Video object referring segmentation involves segmenting specific objects in a video based on language descriptions. It identifies and isolates the referred objects accurately across frames, where MeViS~\bluecite{ding2023mevis} is a common benchmark for this task.
    
    \item \textit{Spatiotemporal Grounding:} Spatiotemporal grounding aims to identify and localize specific objects or events within a video's spatial and temporal dimensions based on a given query. 
    Datasets like {Vid-STG}~\bluecite{zhang2020does}, HC-STVG~\bluecite{hc-stvg-tang2021human}, {Ego4D-MQ and Ego4D-NLQ}~\bluecite{grauman2022ego4d} are proposed to aid the training and testing for this task. 
\end{itemize}

\subsection{Background for LLMs}

Language models are trained to learn a joint probability distribution $p(x_{1:L})$ with a sequence of text tokens $x_{1:L}$. This joint distribution is usually equivalent to a product of the conditional probabilities conditioned on each token with the chain rule:
\begin{equation}
    p(x_{1:L})=\prod_{i=1}^{L}p(x_{i}|x_{1:i-1}),
\end{equation}
where $L$ is the sequence length. 

Large Language Models (LLMs) refer to language models with a large number of parameters, \textit{e.g.}, billions. The architecture of LLMs incorporates a text tokenizer and multiple self-attention layers. LLMs are trained in a teacher-forcing manner to predict the next token's probability, where the generation process utilizes the autoregressive paradigm:
\begin{equation}
    \mathcal{M}(x_{1:i-1}) = p(x_i \mid x_{1:i-1}),
\end{equation}
where $\mathcal{M}$ represents an LLM.

Decoding strategies dictate how to harness the next token probability and select the next token $y_t$ from the set $S$ of all possible tokens in the vocabulary, which includes special tokens such as $<$SOS$>$, $<$EOS$>$, and $<$PAD$>$. Greedy decoding, the simplest strategy, selects the token with the highest probability, formalized as:
\begin{equation}
    x_t = \argmax_{s \in S} \log p_{\mathcal{M}}(s \mid \bm{x}_{1:t-1}).
\end{equation}
Besides deterministic strategies, sampling strategies that randomly select the next tokens using model probability are also popular in real applications. These strategies provide diverse outputs and enable self-consistency methods.

Large language models typically exhibit the following characteristics:
\begin{itemize}
    \item \textit{Scaling Laws}~\bluecite{kaplan2020scaling}: With a significant extension of the model size (number of parameters), the pertaining data size, and computational resources, the performance of the model exhibits a pattern of regular growth, which can help researchers and engineers predict performance improvements or make effective decisions on model design and training.
    \item \textit{Emergent Abilities}~\bluecite{zhao2023survey}: When the parameter size and volume of training data for a large language model exceed a certain magnitude, some novel capabilities emerge, such as in-context learning, instruction following, and step-by-step reasoning.
    \textit{In-context learning} allows the model to learn and make predictions based on the context provided within the input text without requiring explicit retraining. \textit{Instruction following} enables the model to perform tasks based on natural language instructions. \textit{Step-by-step reasoning}, e.g. chain-of-thought (CoT), allows the model to follow a logical sequence of steps to arrive at a conclusion, which is particularly useful for solving complex problems.
\end{itemize}

LLMs possess extensive generalization abilities that can be applied to various downstream tasks, including multi-modal tasks. Multimodal Large Language Models (MLLMs)~\bluecite{zhu2023minigpt, liu2023visual, chen2023shikra, xuan2023pink, hua2024finematch} typically incorporate multimodal encoders, cross-modality aligners, and an LLM core structure. By combining multimodal encoders with the LLM, MLLMs excel at integrating visual and linguistic contexts to produce detailed content.



\section{Vid-LLMs}
\label{sec:model}

In this section, we introduce a novel taxonomy of Vid-LLMs, providing a comprehensive overview of their classification. Following this, we explore the diverse training strategies that empower Vid-LLMs to achieve their capabilities.
\begin{figure}[!h]
    \centering
    \includegraphics[width=0.95\linewidth]{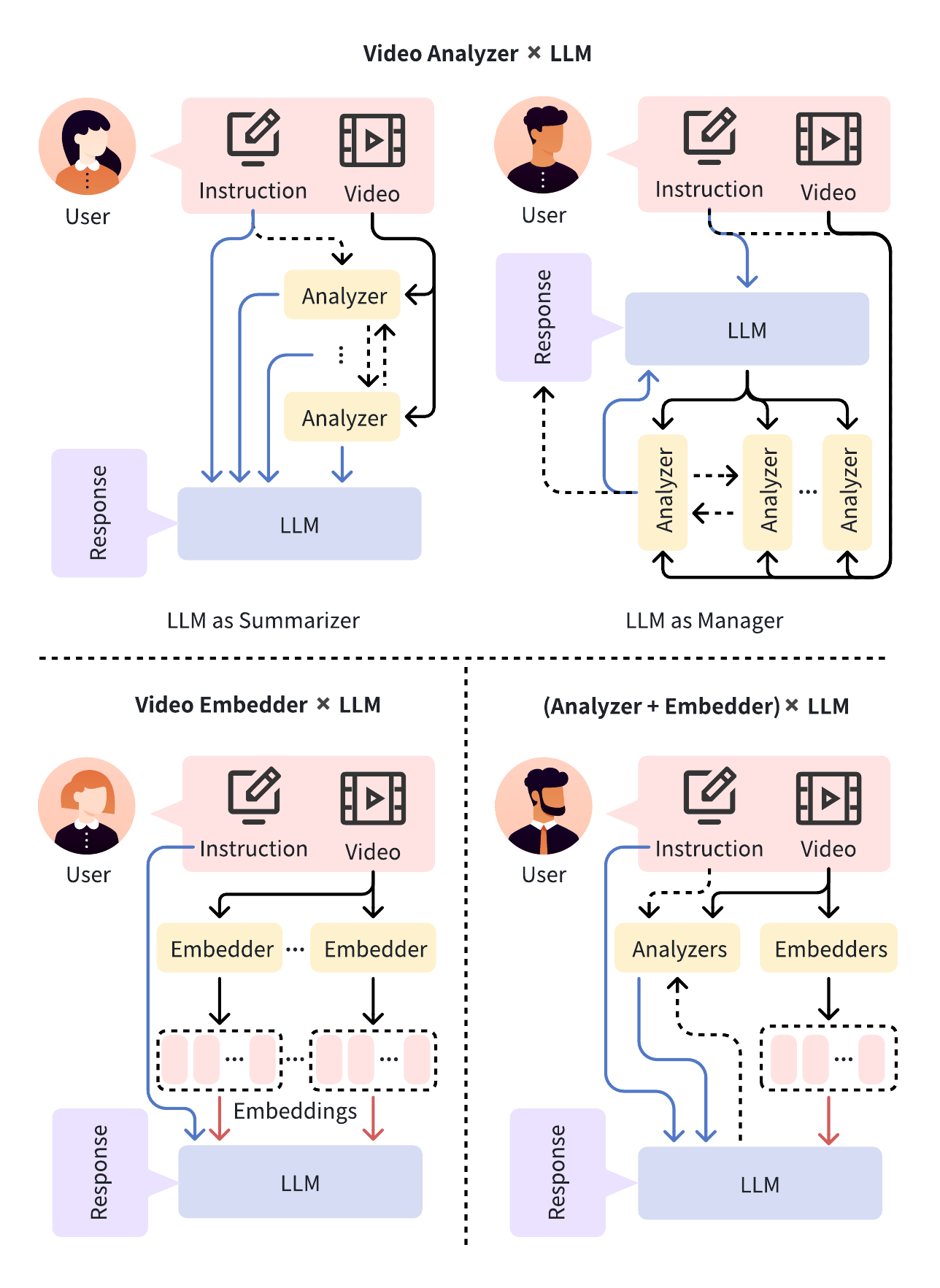}
    \caption{The figure illustrates three primary frameworks for Vid-LLMs: (1) \textit{Video Analyzer $\times$ LLM}, where video analyzers convert video inputs to textual analysis for the LLM; (2) \textit{Video Embedder $\times$ LLM}, where video embedders generate vector representations (embeddings) for the LLM to process; (3) \textit{(Analyzer + Embedder) $\times$ LLM}, a hybrid approach combining both analyzers and embedders to provide the LLM with textual analysis and embeddings. The arrows indicate the direction of information flow, with dashed arrows representing optional paths. \textcolor{st_color}{Blue} arrows denote textual information flow, while \textcolor{t_color}{red} arrows denote embeddings.}
    \label{fig:framework}
\vspace{-1em}
\end{figure}
\tikzstyle{my-box}=[
    rectangle,
    draw=hidden-draw,
    rounded corners,
    text opacity=1,
    minimum height=2em,
    minimum width=5em,
    inner sep=2pt,
    align=center,
    fill opacity=.5,
    line width=0.8pt,
]
\tikzstyle{leaf}=[my-box, minimum height=2em,
     text=black, align=left,font=\normalsize,
    inner xsep=2pt,
    inner ysep=4pt,
    line width=0.8pt,
]
\definecolor{mygreen}{RGB}{209, 215, 239}

\definecolor{mypink}{RGB}{239, 187, 189}
\definecolor{myorange}{RGB}{237, 198, 131}
\definecolor{myblue}{RGB}{60, 105, 251}
\tikzstyle{plus-box}=[my-box, fill=mygreen, draw=mygreen]
\tikzstyle{embedder-box}=[my-box, fill=mypink, draw=mypink]
\tikzstyle{analyzer-box}=[my-box, fill=myorange, draw=myorange]
\tikzstyle{main-box}=[my-box, fill=myblue, draw=myblue]

\begin{figure*}[!th]
    \centering
    \resizebox{0.85\textwidth}{!}{
        \begin{forest}
            forked edges,
            for tree={
                grow=east,
                reversed=true,
                anchor=base west,
                parent anchor=east,
                child anchor=west,
                base=left,
                font=\large,
                rectangle,
                draw=hidden-draw,
                rounded corners,
                align=left,
                minimum width=6em,
                edge+={darkgray, line width=1pt},
                s sep=20pt,
                inner xsep=3pt,
                inner ysep=3pt,
                line width=0.8pt,
                ver/.style={rotate=90, child anchor=north, parent anchor=south, anchor=center},
            },
            where level=1{text width=10.2em,font=\normalsize,}{},
            where level=2{text width=8.6em,font=\normalsize,}{},
            where level=3{text width=8.2em,font=\normalsize,}{},
            where level=4{text width=8.2em,font=\normalsize,}{},
            [
                Vid-LLMs, ver, my-box, text width=12em
                [                
                    Video~Analyzer $\times$ LLM, analyzer-box, edge={myorange}
                    [
                        LLM as Summarizer, analyzer-box, edge={myorange}
                        [
                            GIT-LLaVA~\bluecite{kalarani2024gitllava}{, }
                            \textcolor{t_color}{MM-Screenplayer}~\bluecite{wu2024mmscreenplayer}{, }
                            \textcolor{t_color}{MoReVQA}~\bluecite{min2024morevqa}{, }\\
                            IG-VLM~\bluecite{Kim2024IG-VLM}{, }
                            LangRepo~\bluecite{Kahatapitiya2024LangRepo}{, }
                            MVU~\bluecite{Ranasinghe2024MVU}{, }\\
                            Video ReCap~\bluecite{Islam2024VideoReCap}{, }
                            \textcolor{t_color}{LLoVi}~\bluecite{Zhang2024LLoVi}{, }
                            \textcolor{t_color}{Grounding-Prompter}~\bluecite{Chen2023Grounding-prompter}{, }\\
                            VIDOSC~\bluecite{Xue2024VIDOSC}{, }
                            AntGPT~\bluecite{zhao2023antgpt}{, }
                            VAST~\bluecite{chen2023vast}{, }\\
                            \textcolor{t_color}{VLog}~\bluecite{VLogGithub}{, }
                            LaViLa~\bluecite{zhao2023lavila}{ }
                            , leaf, text width=27em, edge={myorange}, draw=myorange
                        ]
                    ]
                    [
                        LLM as Manager, analyzer-box, edge={myorange}
                        [
                            \textcolor{t_color}{DrVideo}~\bluecite{ma2024drvideo}{, }
                            \textcolor{t_color}{OmAgent}~\bluecite{zhang2024omagent}{, }
                            LVNet~\bluecite{park2024lvnet}{, }\textcolor{t_color}{GPTSee}~\bluecite{Sun2024GPTSee}{, }\\
                            VideoTree~\bluecite{wang2024videotree}{, }
                            \textcolor{t_color}{LAVAD}~\bluecite{zanella2024lavad}{, }
                            TraveLER~\bluecite{shang2024traveler}{, }\\
                            \textcolor{st_color}{RAVA}~\bluecite{Cao2024RAVA}{, }
                            SCHEMA~\bluecite{Niu2024SCHEMA}{, }\textcolor{st_color}{HuggingGPT}~\bluecite{shen2024hugginggpt}{, }\\
                            TV-TREES~\bluecite{S2024TV-TREES}{, }
                            \textcolor{st_color}{VideoAgent}~\bluecite{Fan2024VideoAgent}{, }
                            \textcolor{t_color}{VideoAgent}~\bluecite{Wang2024VideoAgent}{, }\\
                            \textcolor{st_color}{VURF}~\bluecite{Mahmood2024VURF}{, }
                            KEPP~\bluecite{Nagasinghe2024KEPP}{, }
                            \textcolor{st_color}{DoraemonGPT}~\bluecite{Yang2024DoraemonGPT}{, }
                            \textcolor{st_color}{Hawk}~\bluecite{tang2024hawk}{, }\\
                            \textcolor{t_color}{LifelongMemory}~\bluecite{Wang2024LifelongMemory}{, }
                            \textcolor{st_color}{ProViQ}~\bluecite{Choudhury2023ProViQ}{, }
                            \textcolor{st_color}{AssistGPT}~\bluecite{gao2023assistgpt}{, }\\
                            \textcolor{t_color}{Video ChatCaptioner}~\bluecite{wang2023chatvideo}{, }
                            \textcolor{st_color}{ChatVideo}~\bluecite{wang2023chatvideo}{, }
                            \textcolor{st_color}{ViperGPT}~\bluecite{suris2023vipergpt}{ }
                            , leaf, text width=27em, edge={myorange}, draw=myorange
                        ]
                    ]
                ]
                [
                    Video Embedder $\times$ LLM, embedder-box, edge={mypink}
                    [
                        LLM as Text\\Decoder, embedder-box, edge={mypink}
                        [
                            Artemis~\bluecite{qiu2024artemis}{, }
                            EmoLLM~\bluecite{yang2024emollm}{, }
                            FTFV-LLM~\bluecite{chen2024ftfvllm}{, }\\
                            Flash-VStream~\bluecite{zhang2024flashvstream}{, }
                            LLAVIDAL~\bluecite{chakraborty2024llavidal}{, }
                            LongVA~\bluecite{zhang2024longva}{, }\\
                            ShareGPT4Video~\bluecite{chen2024sharegpt4video}{, }
                            VIM~\bluecite{du2024vim}{, }
                            Video-SALMONN~\bluecite{sun2024videosalmonn}{, }\\
                            VideoGPT+~\bluecite{maaz2024videogpt+}{, }
                            VideoLLaMA 2~\bluecite{cheng2024videollama2}{, }
                            MotionLLM~\bluecite{chen2024motionllm}{, }\\
                            VideoChat2~\bluecite{li2024mvbench}{, }
                            Shotluck Holmes~\bluecite{luo2024shotluckholmes}{, }
                            VideoStreaming~\bluecite{qian2024videostreaming}{, }\\
                            VideoNarrator~\bluecite{yang2024videonarrator}{, }
                            TOPA~\bluecite{li2024topa}{, }
                            AutoAD III~\bluecite{han2024autoadiii}{, }GCG~\bluecite{Wang2024GCG}{, }\\
                            LLaVA-Hound-DPO~\bluecite{zhang2024llavahounddpo}{, }
                            RED-VILLM~\bluecite{huang2024redvillm}{, }
                            Koala~\bluecite{tan2024koala}{, }\\
                            LongVLM~\bluecite{weng2024longvlm}{, }
                            MA-LMM~\bluecite{he2024malmm}{, }
                            MiniGPT4-Video~\bluecite{ataallah2024minigpt4video}{, }\\
                            Pegasus-v1~\bluecite{jung2024pegasusv1}{, }
                            PLLaVA~\bluecite{xu2024pllava}{, }
                            ST-LLM~\bluecite{liu2024stllm}{, }COSMO~\bluecite{Wang2024COSMO}{, }\\
                            Tarsier~\bluecite{wang2024tarsier}{, }
                            X-VARS~\bluecite{held2024xvars}{, }
                            CAT~\bluecite{Ye2024CAT}{, }
                            VideoLLM~\bluecite{chen2023videollm}{, }\\
                            InternVideo2~\bluecite{Wang2024InternVideo2}{, }
                            MovieLLM~\bluecite{Song2024MovieLLM}{, }
                            IVAwithLLM~\bluecite{Li2024IVAwithLLM}{, }\\
                            LSTP~\bluecite{Wang2024LSTP}{, }
                            LVCHAT~\bluecite{Wang2024LVCHAT}{, }
                            OSCaR~\bluecite{Nguyen2024OSCaR}{, }Slot-VLM~\bluecite{Xu2024Slot-VLM}{, }\\
                            AV-LLM~\bluecite{Shu2023AV-LLM}{, }
                            Emu2~\bluecite{Sun2024Emu2}{, }
                            MMICT~\bluecite{Chen2023MMICT}{, }
                            VaQuitA~\bluecite{Wang2023VaQuitA}{, }\\
                            VILA~\bluecite{Lin2024VILA}{, }
                            Vista-LLaMA~\bluecite{Ma2023Vista-LLaMA}{, }
                            Chat-UniVi~\bluecite{Jin2024Chat-UniVi}{, }\\
                            LLaMA-VID~\bluecite{li2023llama-vid}{, }
                            Video-LLaVA~\bluecite{lin2023video}{, }
                            LLaMA-VQA~\bluecite{ko2023llamavqa}{, }\\
                            MovieChat~\bluecite{song2023moviechat}{, }
                            LLMVA-GEBC~\bluecite{tang2023llmva}{, }
                            Macaw-LLM~\bluecite{lyu2023macaw}{, }\\
                            VALLEY~\bluecite{luo2023valley}{, }
                            Video-ChatGPT~\bluecite{maaz2023video}{, }
                            Video-LLaMA~\bluecite{zhang2023video}{, }\\
                            mPLUG-video~\bluecite{xu2023youku}{, }
                            ChatBridge~\bluecite{zhao2023chatbridge}{, }
                            Otter~\bluecite{li2023otter}
                            , leaf, text width=28em, edge={mypink}, draw=mypink
                        ]
                    ]
                    [
                        LLM as Regressor, embedder-box, edge={mypink}
                        [
                            \textcolor{t_color}{Holmes-VAD}~\bluecite{zhang2024holmesvad}{, }
                            \textcolor{t_color}{VideoLLM-online}~\bluecite{chen2024videollmonline}{, }
                            \textcolor{st_color}{VLM4HOI}~\bluecite{bansal2024hoiref}{, }\\
                            \textcolor{t_color}{V2Xum-LLaMA}~\bluecite{hua2024v2xum}{, }
                            \textcolor{t_color}{AVicuna}~\bluecite{tang2024avicuna}{, }
                            \textcolor{st_color}{Elysium}~\bluecite{wang2024elysium}{, }\\
                            \textcolor{t_color}{HawkEye}~\bluecite{Wang2024HawkEye}{, }
                            \textcolor{t_color}{LITA}~\bluecite{Huang2024LITA}{, }
                            \textcolor{st_color}{OmniViD}~\bluecite{Wang2024OmniViD}{, }
                            \textcolor{t_color}{SeViLA}~\bluecite{Yu2023SeViLA}{, }\\
                            \textcolor{st_color}{GroundingGPT}~\bluecite{Li2024GroundingGPT}{, }
                            \textcolor{t_color}{TimeChat}~\bluecite{Ren2024TimeChat}{, }
                            \textcolor{t_color}{VTimeLLM}~\bluecite{huang2023vtimellm}
                            , leaf, text width=27em, edge={mypink}, draw=mypink
                        ]
                    ]
                    [
                        LLM as Hidden\\Layer, embedder-box, edge={mypink}
                        [
                            \textcolor{t_color}{VTG-LLM}~\bluecite{guo2024vtgllm}{, }
                            \textcolor{st_color}{VITRON}~\bluecite{fei2024vitron}{, }
                            \textcolor{t_color}{VTG-GPT}~\bluecite{Xu2024VTG-GPT}{, }\\
                            \textcolor{t_color}{Momentor}~\bluecite{Qian2024Momentor}{, }
                            \textcolor{t_color}{VidDetours}~\bluecite{Ashutosh2024VidDetours}{, }
                            OneLLM~\bluecite{Han2023OneLLM}{, }\\
                            GPT4Video~\bluecite{wang2023gpt4video}{ }
                            , leaf, text width=23em, edge={mypink}, draw=mypink
                        ]
                    ]
                ]
                [
                    (Analyzer + Embedder)\\$\times$ LLM, plus-box, edge={mygreen}
                    [
                        LLM as Manager, plus-box, edge={mygreen}
                        [
                            \textcolor{t_color}{MM-VID}~\bluecite{lin2023mmvid}{}
                            , leaf, text width=7em, draw=mygreen, edge={mygreen}
                        ]
                    ]
                    [
                        LLM as Summarizer, plus-box, edge={mygreen}
                        [
                            SUM-shot~\bluecite{Han2023SUM-shot}{}
                            , leaf, text width=7em, draw=mygreen, edge={mygreen}
                        ]
                    ]
                    [
                        LLM as Regressor, plus-box, edge={mygreen}
                        [
                            \textcolor{t_color}{Vriptor}~\bluecite{yang2024vript}{, }
                            \textcolor{st_color}{Merlin}~\bluecite{yu2023merlinempowering}{, }
                            \textcolor{st_color}{VideoChat}~\bluecite{li2023videochat}{, }
                            \textcolor{t_color}{Vid2Seq}~\bluecite{yang2023vid2seq}{ }
                            , leaf, text width=26.5em, draw=mygreen, edge={mygreen}
                        ]
                    ]
                    [
                        LLM as Text\\Decoder, plus-box, edge={mygreen}
                        [
                            \textcolor{t_color}{Uni-AD}~\bluecite{Wang2024Uni-AD}{, }
                            \textcolor{t_color}{MM-narrator}~\bluecite{zhang2023mm-narrator}{, }
                            Vamos~\bluecite{Wang2024Vamos}{, }\\
                            Auto-AD II~\bluecite{han2023autoad}{, }
                            \textcolor{t_color}{CAT-V}~\bluecite{tang2025caption}{ }
                            , leaf, text width=22em, draw=mygreen, edge={mygreen}
                        ]
                    ]
                    [
                        LLM as Hidden\\Layer, plus-box, edge={mygreen}
                        [
                            \textcolor{st_color}{PG-Video-LLaVA}~\bluecite{munasinghe2023pg}{ }
                            , leaf, text width=10em, draw=mygreen, edge={mygreen}
                        ]
                    ]
                ]
            ]
        \end{forest}
                                                                           }
    \caption{Taxonomy of Video Understanding with Large Language Models (Vid-LLMs), consists of \textit{Video Analyzer $\times$ LLM}, \textit{Video Embedder $\times$ LLM} and \textit{(Analyzer + Embedder) $\times$ LLM}, and the sub-categories are \textit{LLM as Summarizer/Manager/Text Decoder/Regressor/Hidden Layer}. Font color indicates the granularity of video understanding supported by the Vid-LLMs: black for abstract understanding, \textcolor{t_color}{red} for temporal understanding, and \textcolor{st_color}{blue} for spatiotemporal understanding.}
    \label{fig:taxonomy}
\end{figure*}
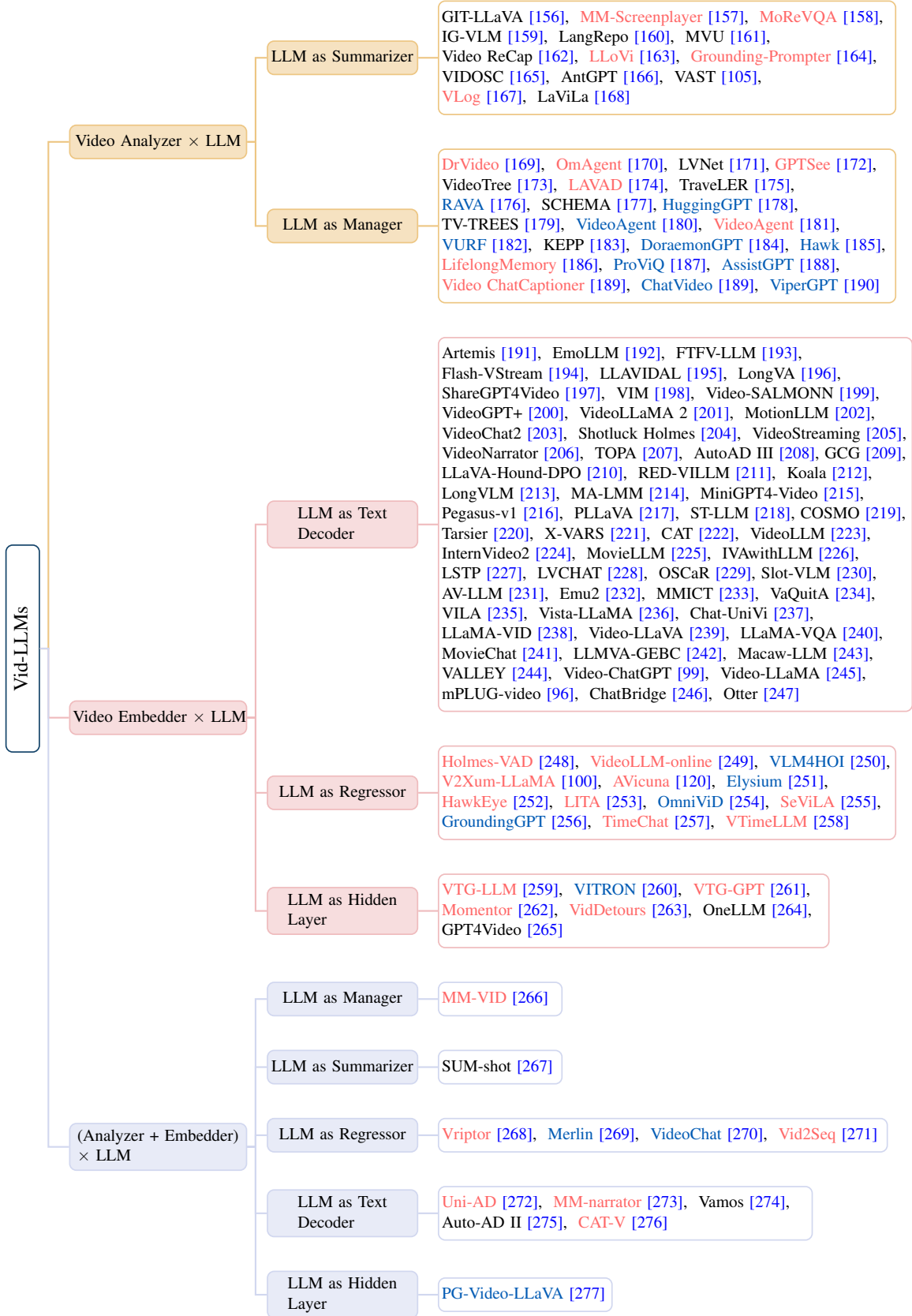
\subsection{Taxonomy}
Based on the method of processing input video, we categorize Vid-LLMs into three primary types: \textit{Video Analyzer $\times$ LLM}, \textit{Video Embedder $\times$ LLM}, and \textit{(Analyzer + Embedder) $\times$ LLM}. Each category represents a unique approach to integrating video processing with LLMs, as illustrated in \Cref{fig:framework}.

\subsubsection{Video Analyzer $\times$ LLM}
 Video Analyzer is defined as a module that takes video input and outputs an analysis of the video, typically in text form, which facilitates LLM processing. This text may include video captions, dense video captions (detailed descriptions of all events in the video with timestamps), object tracking results (labels, IDs, and bounding boxes of objects), as well as transcripts of other modalities present in the video, such as speech recognition results from ASR or subtitle recognition results from OCR. The text generated by the Video Analyzer can be directly fed into the subsequent LLM, inserted into pre-prepared templates before being fed into the LLM, or converted into a temporary database format for the LLM to retrieve later.

\begin{figure*}[!h]
    \centering
    \includegraphics[width=0.95\linewidth]{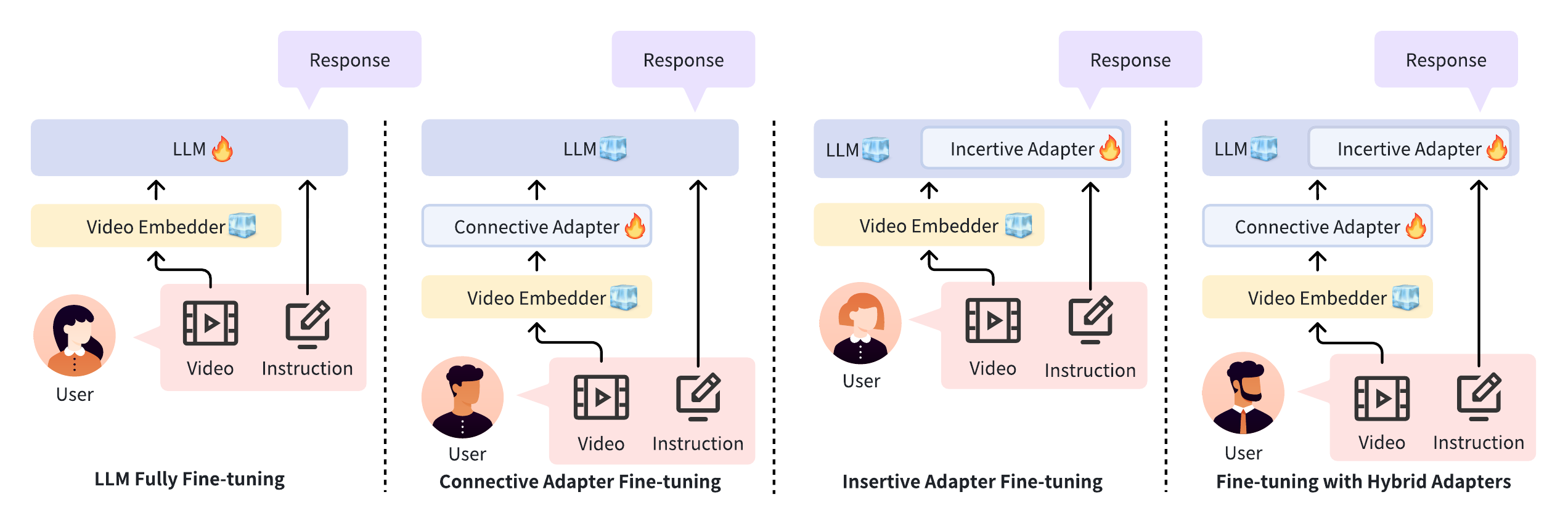}
    \caption{Four types of Vid-LLMs fine-tuning strategies,  classified according to their specific training methods: LLM fully fine-tuning, fine-tuning with connective adapters, insertive adapters, and hybrid methods.}
    \label{fig:finetuning}
\vspace{-1em}
\end{figure*}

For the \textit{Video Analyzer $\times$ LLM} category, we further create two subcategories, \textit{LLM as Summarizer} and \textit{LLM as Manager}, based on the function of the LLM within the Vid-LLM system:

\begin{itemize}
    \item \textit{LLM as Summarizer:} In this subcategory, the primary function of the LLM is to summarize the analysis obtained from the Video Analyzers. The summarization approach varies based on the prompts provided to the LLM, ranging from highly condensed summary texts and captions to comprehensive summaries for answering specific questions. Notably, in \textit{Video Analyzer $\times$ LLM as Summarizer} systems, the information flow is usually unidirectional (see \Cref{fig:framework}), with data flowing from the video to the Video Analyzer and then to the LLM, without any reverse process. Examples of Vid-LLMs in the \textit{Video Analyzer $\times$ LLM as Summarizer} category include: LaViLa~\bluecite{zhao2023lavila}, VLog~\bluecite{VLogGithub}, VAST~\bluecite{chen2023vast}, AntGPT~\bluecite{zhao2023antgpt},  VIDOSC~\bluecite{Xue2024VIDOSC}, Grounding-Prompter~\bluecite{Chen2023Grounding-prompter}, LLoVi~\bluecite{Zhang2024LLoVi}, Video ReCap~\bluecite{Islam2024VideoReCap}, MVU~\bluecite{Ranasinghe2024MVU}, LangRepo~\bluecite{Kahatapitiya2024LangRepo}, IG-VLM~\bluecite{Kim2024IG-VLM}, MoReVQA~\bluecite{min2024morevqa}, MM-Screenplayer~\bluecite{wu2024mmscreenplayer}, GIT-LLaVA~\bluecite{kalarani2024gitllava}, etc.


    \item \textit{LLM as Manager:} In this subcategory, the LLM primarily coordinates the overall system's operation. It may actively generate commands to invoke various Video Analyzers to produce the desired results for the user, call the Video Analyzer and further process the obtained analysis before returning it to the user, or engage in multiple rounds of interaction with the Video Analyzer. Compared to the \textit{LLM as Summarizer category}, the \textit{LLM as Manager} category is more flexible and can be distinguished by its information flow complexity. Examples of Vid-LLMs in the \textit{Video Analyzer $\times$ LLM as Manager} category include: ViperGPT~\bluecite{suris2023vipergpt}, Video ChatCaptioner~\bluecite{chen2023videochatcap}, ChatVideo~\bluecite{wang2023chatvideo}, AssistGPT~\bluecite{gao2023assistgpt}, HuggingGPT~\bluecite{shen2024hugginggpt}, {Hawk}~\bluecite{tang2024hawk}, ProViQ~\bluecite{Choudhury2023ProViQ}, LifelongMemory~\bluecite{Wang2024LifelongMemory}, DoraemonGPT~\bluecite{Yang2024DoraemonGPT}, KEPP~\bluecite{Nagasinghe2024KEPP}, VURF~\bluecite{Mahmood2024VURF}, VideoAgent (by Stanford)~\bluecite{Wang2024VideoAgent}, VideoAgent (by PKU)~\bluecite{Fan2024VideoAgent}, TV-TREES~\bluecite{S2024TV-TREES}, SCHEMA~\bluecite{Niu2024SCHEMA}, RAVA~\bluecite{Cao2024RAVA}, GPTSee~\bluecite{Sun2024GPTSee}, TraveLER~\bluecite{shang2024traveler}, LAVAD~\bluecite{zanella2024lavad}, VideoTree~\bluecite{wang2024videotree}, LVNet~\bluecite{park2024lvnet}, OmAgent~\bluecite{zhang2024omagent}, DrVideo~\bluecite{ma2024drvideo}, etc.

\end{itemize}

\subsubsection{Video Embedder $\times$ LLM}
A Video Embedder typically refers to a visual backbone/video encoder, such as ViT or CLIP, used to convert input videos into vector representations, known as video embeddings or video tokens. Some Embedders encode other modalities within the video, such as audio (e.g., CLAP~\bluecite{elizalde2023clap}), which are also categorized under Video Embedder here (note that we do not consider the LLM's tokenizer as an embedder). Unlike the text generated by the Video Analyzer, the vectors generated by the Video Embedder cannot be directly utilized by the LLM and usually require an adapter to map these embeddings from the vision's (or other modalities') semantic space to the text semantic space of the LLM's input tokens.

For the \textit{Video Embedder $\times$ LLM} category, we also classify them into subcategories based on the LLM's function within the Vid-LLM system:

\begin{itemize}
    \item \textit{LLM as Text Decoder:} In this subcategory, the LLM receives embeddings from the Video Embedder as input and decodes them into text outputs based on prompts or instructions. These tasks generally do not require fine-grained understanding or precise spatiotemporal localization, focusing mainly on general QA or captioning. Thus, the Vid-LLM behaves like a standard LLM during decoding. Examples of Vid-LLMs in the \textit{Video Embedder $\times$ LLM as Text Decoder} category include: VideoLLM~\bluecite{chen2023videollm}, Otter~\bluecite{li2023otter}, Video-LLaMA~\bluecite{zhang2023video}, Video-ChatGPT~\bluecite{maaz2023video}, VALLEY~\bluecite{luo2023valley}, Macaw-LLM~\bluecite{lyu2023macaw}, MovieChat~\bluecite{song2023moviechat}, Video-LLaVA~\bluecite{lin2023video}, Chat-UniVi~\bluecite{Jin2024Chat-UniVi}, Vista-LLaMA~\bluecite{Ma2023Vista-LLaMA}, VILA~\bluecite{Lin2024VILA}, GPT4Video~\bluecite{wang2023gpt4video}, MovieLLM~\bluecite{Song2024MovieLLM}, InternVideo2~\bluecite{Wang2024InternVideo2}, MiniGPT4-Video~\bluecite{ataallah2024minigpt4video}, VideoChat2~\bluecite{li2024mvbench}, VideoLLaMA 2~\bluecite{cheng2024videollama2}, etc. See \Cref{fig:taxonomy} for the complete list.
    \item \textit{LLM as Regressor:} In this subcategory, the LLM receives embeddings from the Video Embedder as input and, like the Text Decoder, can output text. However, unlike the Text Decoder, the \textit{LLM as Regressor} can also predict continuous values, such as timestamp localization in videos and bounding box coordinates for object trajectories, functioning similarly to a regressor performing regression tasks, even though it is fundamentally performing classification. Examples of Vid-LLMs in the \textit{Video Embedder $\times$ LLM as Regressor} category include:  VTimeLLM~\bluecite{huang2023vtimellm}, SeViLA~\bluecite{Yu2023SeViLA}, TimeChat~\bluecite{Ren2024TimeChat},  GroundingGPT~\bluecite{Li2024GroundingGPT}, OmniViD~\bluecite{Wang2024OmniViD}, LITA~\bluecite{Huang2024LITA}, HawkEye~\bluecite{Wang2024HawkEye}, Elysium~\bluecite{wang2024elysium}, AVicuna~\bluecite{tang2024avicuna}, V2Xum-LLaMA~\bluecite{hua2024v2xum}, VLM4HOI~\bluecite{bansal2024hoiref}, VideoLLM-online~\bluecite{chen2024videollmonline}, Holmes-VAD~\bluecite{zhang2024holmesvad}, etc.

    \item \textit{LLM as Hidden Layer:} In this subcategory, the LLM also receives video embeddings as input but does not directly output text. Instead, it connects to a specially designed task-specific head to perform actual regression tasks, such as event time localization or object bounding box prediction in videos, while maintaining the LLM's text output capability. Examples of Vid-LLMs in the \textit{Video Embedder $\times$ LLM as Hidden Layer} category include:  GPT4Video~\bluecite{wang2023gpt4video}, OneLLM~\bluecite{Han2023OneLLM}, VidDetours~\bluecite{Ashutosh2024VidDetours}, Momentor~\bluecite{Qian2024Momentor}, VTG-GPT~\bluecite{Xu2024VTG-GPT}, VITRON~\bluecite{fei2024vitron}, VTG-LLM~\bluecite{guo2024vtgllm}, etc.

\end{itemize}

\subsubsection{(Analyzer + Embedder) $\times$ LLM}
This category of Vid-LLMs is relatively rare. As the name suggests, it involves simultaneously using a Video Analyzer to obtain textual analysis of the video and a Video Embedder to encode the video into embeddings. The LLM receives both types of inputs along with other prompts/instructions and outputs responses to complete tasks. The subcategories here can flexibly be any of the Summarizer/Manager/Text Decoder/Regressor/Hidden Layer categories. Vid-LLMs in the \textit{(Analyzer + Embedder) $\times$ LLM} category include: Vid2Seq~\bluecite{yang2023vid2seq}, VideoChat~\bluecite{li2023videochat},  MM-VID~\bluecite{lin2023mmvid}, Auto-AD II~\bluecite{han2023autoad}, Vamos~\bluecite{Wang2024Vamos}, PG-Video-LLaVA~\bluecite{munasinghe2023pg}, MM-Narrator~\bluecite{zhang2023mm-narrator}, SUM-shot~\bluecite{Han2023SUM-shot}, Merlin~\bluecite{yu2023merlinempowering}, Uni-AD~\bluecite{Wang2024Uni-AD}, Vriptor~\bluecite{yang2024vript}, etc.

\subsection{Training Strategies for Vid-LLMs}

\subsubsection{Training-free Vid-LLMs}
Many Vid-LLMs systems are built on powerful LLMs with strong zero-shot, in-context learning, and Chain-of-Thought capabilities. These systems do not require training in the parameters of the LLM or other modules. Most Vid-LLMs in the \textit{Video Analyzer $\times$ LLM} category are training-free because the information from the video and other accompanying modalities has already been parsed into text. At this point, the video understanding task has been transformed into a text understanding task. Since LLMs can unify almost all NLP tasks into generation tasks, they can also handle many video understanding tasks. SlowFast-LLaVA 
\bluecite{xu2024slowfastllava} is a training-free Video LLM that uses a two-stream input design to capture both spatial semantics and temporal context without fine-tuning and demonstrates capabilities across various video understanding benchmarks.

\begin{table*}[]
\centering
\caption{Comparison of video understanding models with large language models, sorted by their release dates. The table presents key details for each method, including the number of training frames, video embedders, utilization of audio information, model adaptation approaches, computational resources, the specific large language model employed, and the corresponding number of parameters. Entries marked with a hyphen (``-") indicate undisclosed details in the respective papers.
}
\label{tab:model_card}
\setlength{\tabcolsep}{1pt}
\resizebox{\textwidth}{!}{%
\begin{tabular}{l|ccccccccc}
\hline
\multicolumn{1}{c|}{\textbf{Model}} & \textbf{\#Frame} & \textbf{Video Embedder} & \textbf{Sound} & \textbf{Speech} & \textbf{Adapter} & \textbf{Hardware} & \textbf{LLM} & \textbf{LLM Size} & \textbf{Date} \\ \hline
Socratic Models~\bluecite{zeng2022socratic} & Varying & CLIP ViT-L/14 & \xmark & \cmark & - & 1 V100 GPU & RoBERTa, GPT-3 & - & 05/2022 \\
\rowcolor[HTML]{ECF4FF} 
LaViLa~\bluecite{zhao2023lavila} & 4 & TimeSformer-B/L & \xmark & \xmark & Cross-Attention & 32 V100 GPUs & GPT-2 XL frozeen & - & 12/2022 \\
Vid2Seq~\bluecite{yang2023vid2seq} & 100 & CLIP ViT-L/14 & \xmark & \cmark & Transformer Encoder & 64 TPU v4 & T5 & 0.2B & 02/2023 \\
\rowcolor[HTML]{ECF4FF} 
ViperGPT~\bluecite{suris2023vipergpt} & - & - & \xmark & \xmark & - & - & - & - & 03/2023 \\
Vid ChatCaptioner~\bluecite{chen2023videochatcap} & 100 & - & \xmark & \xmark & BLIP-2 & 24G GPU & ChatGPT & 20B & 04/2023 \\
\rowcolor[HTML]{ECF4FF} 
VLog~\bluecite{VLogGithub} & - & CLIP ViT-G & \cmark & \cmark & - & - & ChatGPT & 20B & 04/2023 \\
ChatVideo~\bluecite{wang2023chatvideo} & - & - & \cmark & \cmark & - & - & ChatGPT & 20B & 04/2023 \\
\rowcolor[HTML]{ECF4FF} 
VideoChat~\bluecite{li2023videochat} & 4-32 & ViT-G & \xmark & \cmark & MLP+Q-fomer & 1 A10 GPU & StableVicuna & 7B & 05/2023 \\
ChatBridge~\bluecite{zhao2023chatbridge} & 4 & ViT-G & \cmark & \xmark & Perceiver & 8 A100 GPUs & GPT-4 & - & 05/2023 \\
\rowcolor[HTML]{ECF4FF} 
VAST~\bluecite{chen2023vast} & 4,8 & EVA-CLIP, BEATs, BERT-B & \cmark & \cmark & - & 64 V100 GPUs & Vicuna & 13B & 05/2023 \\
Otter~\bluecite{li2023otter} & 4-8 & CLIP ViT-L/14 & \xmark & \xmark & - & 4 RTX-3090 GPUs & LLaMA & 7B & 05/2023 \\
\rowcolor[HTML]{ECF4FF} 
VideoLLM~\bluecite{chen2023videollm} & Varying & \scriptsize 7 Task-Specific Video Encoders & \xmark & \xmark & Linear Layer & - & GPT-2/T5/OPT/LLaMA & \tiny1.5/6.5/6.7/7B & 05/2023 \\
AssistGPT~\bluecite{gao2023assistgpt} & 1/3 FPS & - & \xmark & \cmark & - & 4 A5000 GPUs & Vicuna & 7B & 06/2023 \\
\rowcolor[HTML]{ECF4FF} 
Video-LLaMA~\bluecite{zhang2023video} & 8 & CLIP ViT-G & \cmark & \xmark & Video Q-former & - & Vicuna & 7B & 06/2023 \\
Video-ChatGPT~\bluecite{maaz2023video} & 100 & CLIP ViT-L/14 & \cmark & \cmark & Linear Layer & 8 A100 GPUs & Vicuna-v1.1 & 7B & 06/2023 \\
\rowcolor[HTML]{ECF4FF} 
LLMVA-GEBC~\bluecite{tang2023llmva} & 96 & CLIP ViT-G & \xmark & \xmark & Video Q-former & 2 A6000 GPUs & OPT & 13B & 06/2023 \\
mPLUG-video~\bluecite{xu2023youku} & 8 & TimeSformer & \xmark & \xmark & - & - & GPT/Blood & - & 06/2023 \\
\rowcolor[HTML]{ECF4FF} 
VALLEY~\bluecite{luo2023valley} & Varying & CLIP ViT-L/14 & \xmark & \xmark & Projection Layer & 8 A100 GPUs & StableVicuna & 7B/13B & 06/2023 \\
Macaw-LLM~\bluecite{lyu2023macaw} & Varying & CLIP VIT-B/16 & \xmark & \cmark & Alignment and Integration & - & LLaMA/Vicuna/Bloom & 7B & 06/2023 \\
\rowcolor[HTML]{ECF4FF} 
AntGPT~\bluecite{zhao2023antgpt} & 4 & CLIP ViT-L/14 & \xmark & \xmark & - & A6000 GPUs & Llama2 & 7B & 07/2023 \\
MovieChat~\bluecite{song2023moviechat} & 2048 & ViT-G/14, EVA-CLIP & \xmark & \xmark & Q-former+LSTM & - & GPT-3.5/Claude & - & 07/2023 \\
\rowcolor[HTML]{ECF4FF} 
FAVOR~\bluecite{sun2023finegrained} & Varying & InstructBLIP ViT-G/14 & \cmark & \cmark & Causal Q-Former & - & Vicuna & 7B/13B & 10/2023 \\
Auto-AD II~\bluecite{han2023autoad} & Varying & CLIP ViT-B/32 & \xmark & \xmark & - & - & GPT-2 & - & 10/2023 \\
\rowcolor[HTML]{ECF4FF} 
LLaMA-VQA~\bluecite{ko2023llamavqa} & - & CLIP ViT-L/14 & \xmark & \xmark & Linear Layer & 8 A6000 GPUs & LLaMA & 7B & 10/2023 \\
MM-VID~\bluecite{lin2023mmvid} & Varying & - & \xmark & \cmark & - & - & GPT4 & - & 10/2023 \\
\rowcolor[HTML]{ECF4FF} 
Video-LLaVA~\bluecite{lin2023video} & 8-32 & LanguageBind & \cmark & \xmark & MLP Projection Layer & - & Vicuna/LLaVA & 7B/13B & 11/2023 \\
PG-Video-LLaVA~\bluecite{munasinghe2023pg} & Varying & CLIP ViT-L/14 & \xmark & \cmark & MLP Projection Layer & 4 A100 GPUs & Vicuna/LLaVA & 7B/13B & 11/2023 \\
\rowcolor[HTML]{ECF4FF} 
SeViLA~\bluecite{Yu2023SeViLA} & 32 & BLIP-2 ViT & \xmark & \xmark & Q-former & 4 48GB A6000 GPUs & MiniGPT4 & - & 11/2023 \\
Vamos~\bluecite{Wang2024Vamos} & Varying & CLIP ViT-L/14, BLIP-2 ViT & \xmark & \xmark & LORA / LLaMA-Adapter & A6000 GPU & LLaMA/LLaMA2 & 7B & 11/2023 \\
\rowcolor[HTML]{ECF4FF} 
VTimeLLM~\bluecite{huang2023vtimellm} & 100 & CLIP ViT-L/14 & \xmark & \xmark & Linear Layer & 1 RTX-4090 GPU & Vicuna & 7B/13B & 11/2023 \\
Chat-UniVi~\bluecite{Jin2024Chat-UniVi} & 64 & ViT-L/14, ViT-G & \xmark & \xmark & Spatial Temporal Merging & - & Vicuna & 7B & 11/2023 \\
\rowcolor[HTML]{ECF4FF} 
MM-narrator~\bluecite{zhang2023mm-narrator} & Varying & CLIP ViT & \xmark & \cmark & - & - & GPT-4 & - & 11/2023 \\
GPT4Video~\bluecite{wang2023gpt4video} & Varying & CLIP ViT-L/14 & \xmark & \xmark & Cross-Attention & 8 A100 GPUs & LLaMA & 7B & 11/2023 \\
\rowcolor[HTML]{ECF4FF} 
LLaMA-VID~\bluecite{li2023llama-vid} & 1fps & CLIP ViT-L/14 & \xmark & \xmark & Projector & 8 A100 GPUs & Vicuna & 7B/13B & 11/2023 \\
LLoVi~\bluecite{Zhang2024LLoVi} & 0.5fps & CLIP, TimeSformer, BLIP & \xmark & \xmark & - & 32 RTX 3090 GPUs & GPT-4 & - & 12/2023 \\
\rowcolor[HTML]{ECF4FF} 
AV-LLM~\bluecite{Shu2023AV-LLM} & 32 & CLIP ViT-L/14 & \xmark & \xmark & Linear Projectors & 8×A100 GPUs & Vicuna & 7B & 12/2023 \\
Emu2~\bluecite{Sun2024Emu2} & Varying & EVA-02-CLIP-E-plus & \xmark & \xmark & Linear Projection & - & Emu2-Chat & 38B & 12/2023 \\
\rowcolor[HTML]{ECF4FF} 
Grounding-Prompter~\bluecite{Chen2023Grounding-prompter} & Varying & BLIP & \xmark & \cmark & - & - & GPT-3.5-turbo-16k & - & 12/2023 \\
VIDOSC~\bluecite{Xue2024VIDOSC} & 1fps & CLIP & \xmark & \xmark & - & - & GPT4 & - & 12/2023 \\
\rowcolor[HTML]{ECF4FF} 
LifelongMemory~\bluecite{Wang2024LifelongMemory} & 360 & CLIP ViT-L/14, TimeSformer & \xmark & \xmark & - & - & GPT-4 & - & 12/2023 \\
Merlin~\bluecite{yu2023merlinempowering} & 8 & CLIP ViT-L/14 & \xmark & \xmark & 2D Conv + Linear Layer & - & Vicuna v1.5 & 7B & 12/2023 \\
\rowcolor[HTML]{ECF4FF} 
MMICT~\bluecite{Chen2023MMICT} & 16 & BLIP-2 ViT-G/14 & \xmark & \xmark & Transformers + Linear Projection & 4 V100 GPUs & FlanT5XL, OPT & 2.7B & 12/2023 \\
OneLLM~\bluecite{Han2023OneLLM} & - & CLIP ViT-L/14 & \xmark & \xmark & K Transformer Projectors + 1 MLP & 16 A100 GPUs & LLaMA2 & 7B & 12/2023 \\
\rowcolor[HTML]{ECF4FF} 
SUM-shot~\bluecite{Han2023SUM-shot} & 4 & EVA-CLIP ViT-G/14 & \cmark & \cmark & Q-former & 16 A100-80G GPUs & Vicuna-v0 & 7B & 12/2023 \\
TimeChat~\bluecite{Ren2024TimeChat} & 96 & BLIP-2 ViT-G/14 & \xmark & \xmark & Video Q-Former+Linear Layer+LoRA & - & LLaMA-2 & 7B & 12/2023 \\
\rowcolor[HTML]{ECF4FF} 
VaQuitA~\bluecite{Wang2023VaQuitA} & 100 & CLIP ViT-L/14 & \xmark & \xmark & Video Perceiver + VQ-Former & 8 A100 80GB GPUs & Llama2 & 7B & 12/2023 \\
VILA~\bluecite{Lin2024VILA} & 8 & CLIP ViT-L/14 & \xmark & \xmark & Linear Projector & 16 A100 GPUs & Llama-2 & 7B/13B & 12/2023 \\
\rowcolor[HTML]{ECF4FF} 
Vista-LLaMA~\bluecite{Ma2023Vista-LLaMA} & - & EVA-CLIP, CLIP ViT-L/14 & \xmark & \xmark & Sequential Q-Former with linear layer & 8 A100 80GB GPUs & Vicuna & 7B & 12/2023 \\
ProViQ~\bluecite{Choudhury2023ProViQ} & - & TimeSformer & \xmark & \cmark & - & one A100 GPU & gpt-3.5-turbo & - & 12/2023 \\
\rowcolor[HTML]{ECF4FF} 
COSMO~\bluecite{Wang2024COSMO} & 3 & CLIP ViT-L/14 & \xmark & \xmark & Gated Cross-Attention Layers & 128 V100 GPUs & OPT-IML/RedPajama/Mistral & 1.8/3/7B & 01/2024 \\
VidDetours~\bluecite{Ashutosh2024VidDetours} & - & InternVideo & \xmark & \xmark & Linear Layer & 8 A100 GPUs & LLaMA-2 & 13B & 01/2024 \\
\rowcolor[HTML]{ECF4FF} 
DoraemonGPT~\bluecite{Yang2024DoraemonGPT} & - & - & \xmark & \cmark & - & - & GPT-3.5-turbo & - & 01/2024 \\
GroundingGPT~\bluecite{Li2024GroundingGPT} & 64 & CLIP ViT-L/14 & \xmark & \xmark & Q-former, MLP & 8 A100 GPUs & Vicuna-v1.5 & 7B & 01/2024 \\
\rowcolor[HTML]{ECF4FF} 
GCG~\bluecite{Wang2024GCG} & 4 & EVA-CLIP & \xmark & \xmark & Q-former & - & IntructBLIP-Vicuna & 7B & 01/2024 \\
IVAwithLLM~\bluecite{Li2024IVAwithLLM} & - & no mention & \xmark & \xmark & Linear Layer & 8 A100 GPUs & Vicuna & 7B & 02/2024 \\
\rowcolor[HTML]{ECF4FF} 
LSTP~\bluecite{Wang2024LSTP} & Varying & BLIP-2 ViT-G/14 & \xmark & \xmark & Q-former & 1 A100 GPU & Vicuna & 7B & 02/2024 \\
LVCHAT~\bluecite{Wang2024LVCHAT} & Varying & UMT-L & \xmark & \xmark & QFormer + Linear Layer & 4 A6000 GPUs & Vicuna & 7B & 02/2024 \\
\rowcolor[HTML]{ECF4FF} 
Momentor~\bluecite{Qian2024Momentor} & 300 & CLIP ViT-L/14 & \xmark & \xmark & Projection Layer & 8 A1000 GPUs & LLaMA & 7B & 02/2024 \\
OSCaR~\bluecite{Nguyen2024OSCaR} & - & CLIP ViT-L/14 & \xmark & \xmark & - & - & Vicuna & 7/13B & 02/2024 \\
\rowcolor[HTML]{ECF4FF} 
Slot-VLM~\bluecite{Xu2024Slot-VLM} & 1 & CLIP ViT-L/14 & \xmark & \xmark & Linear Layer & 1 A100 GPU & Vicuna & 7B & 02/2024 \\
Video ReCap~\bluecite{Islam2024VideoReCap} & 4 & TimeSformer & \xmark & \xmark & - & 8 V100 GPUs & GPT-2 & 1.5B & 02/2024 \\
\rowcolor[HTML]{ECF4FF} 
MVU~\bluecite{Ranasinghe2024MVU} & 16 & - & \xmark & \xmark & - & 2 A5000 GPUs & Mistral/LLaMA-2/Gemma & 7B/13B & 03/2024 \\
VideoAgent~\bluecite{Fan2024VideoAgent} & Varying & ViCLIP & \xmark & \xmark & - & - & GPT-4 & - & 03/2024 \\
\rowcolor[HTML]{ECF4FF} 
VideoAgent~\bluecite{Wang2024VideoAgent} & 1 FPS & EVA-CLIP-8B-plus & \xmark & \xmark & - & 1 A6000 GPU & GPT-4 & - & 03/2024 \\
VTG-GP~\bluecite{Xu2024VTG-GPT} & 0.5 FPS & EVA-CLIP & \xmark & \xmark & - & 8 RTX-3090 GPUs & LLaMA-2 & 7B & 03/2024 \\
\rowcolor[HTML]{ECF4FF} 
VURF~\bluecite{Mahmood2024VURF} & - & - & \xmark & \xmark & - & - & GPT-3.5 & - & 03/2024 \\
KEPP~\bluecite{Nagasinghe2024KEPP} & - & - & \xmark & \xmark & - & 2 A100 GPUs & LLaMA-2 & 13B/70B & 03/2024 \\
\rowcolor[HTML]{ECF4FF} 
GPTSee~\bluecite{Sun2024GPTSee} & - & CLIP ViT-B/32, SlowFast & \xmark & \xmark & Cross-attention & 8 RTX 3090 GPUs & - & - & 03/2024 \\
HawkEye~\bluecite{Wang2024HawkEye} & Varying & UMT-L & \xmark & \cmark & Q-former + LORA & 8 V100 GPUs & GPT-4 & - & 03/2024 \\
\rowcolor[HTML]{ECF4FF} 
InternVideo2~\bluecite{Wang2024InternVideo2} & 8 & InternVL-6B+VideoMAE-g & \cmark & \cmark & Q-former + LoRA & 256 A100 GPUs & Mixer-7B + BERT-large & 7B & 03/2024 \\
LangRepo~\bluecite{Kahatapitiya2024LangRepo} & - & - & \xmark & \xmark & - & - & Mistral & 7B/12B & 03/2024 \\
\rowcolor[HTML]{ECF4FF} 
LITA~\bluecite{Huang2024LITA} & 100 & CLIP ViT-L/14 & \xmark & \xmark & MLP + SlowFast Token Pooling & 8 A100 GPUs & Vicuna & 7B/13B & 03/2024 \\
MovieLLM~\bluecite{Song2024MovieLLM} & 1 FPS & CLIP ViT-L/14 & \xmark & \xmark & - & 4 A100 GPUs & Vicuna & 7B & 03/2024 \\
\rowcolor[HTML]{ECF4FF} 
OmniViD~\bluecite{Wang2024OmniViD} & 32 & VideoSwin & \xmark & \xmark & Mixed Q-former+Visual Translator & 8 A100 GPUs & GPT-4 & - & 03/2024 \\
RAVA~\bluecite{Cao2024RAVA} & 30 FPS & - & \xmark & \xmark & - & - & GPT-4 & - & 03/2024 \\
\rowcolor[HTML]{ECF4FF} 
SCHEMA~\bluecite{Niu2024SCHEMA} & - & CLIP ViT-L/14 & \xmark & \xmark & - & 1 V100 GPU & GPT-3.5 & - & 03/2024 \\
TV-TREES~\bluecite{S2024TV-TREES} & 2 FPS & CLIP ViT-L/14 & \xmark & \xmark & - & - & GPT-3.5 & - & 03/2024 \\
\rowcolor[HTML]{ECF4FF} 
IG-VLM~\bluecite{Kim2024IG-VLM} & 6 & CLIP ViT-L/14 & \xmark & \xmark & - & - & LLaVA-v1.6, and GPT-4V & 7B/13B/34B & 03/2024 \\
AVicuna~\bluecite{tang2024avicuna} & 100 & CLIP ViT & \cmark & \xmark & MLP, LoRA & 1 A6000 GPU & Vicuna-v1.5 & 7B & 03/2024 \\
\rowcolor[HTML]{ECF4FF} 
CAT~\bluecite{Ye2024CAT} & - & ImageBind & \cmark & \xmark & Linear Projector & 1 A100 GPU & LLaMA-2 & 7B & 03/2024 \\
Uni-AD~\bluecite{Wang2024Uni-AD} & - & CLIP ViT-B/32 & \cmark & \xmark & Video mapping network & 8 A100 GPUs & LLaMA-2 & 7B & 03/2024 \\
\rowcolor[HTML]{ECF4FF} 
Elysium~\bluecite{wang2024elysium} & Varying & CLIP-ViT-L & \xmark & \xmark & T-Selector & 24 A100-80G & Vicuna & - & 03/2024 \\ \hline
\end{tabular}%
}
\end{table*}

\begin{table*}[]
\centering
\caption{Contiune of \Cref{tab:model_card}.}
\label{tab:model_card2}
\setlength{\tabcolsep}{3pt}
\resizebox{\textwidth}{!}{%
\begin{tabular}{l|ccccccccc}
\hline
\multicolumn{1}{c|}{\textbf{Model}} & \textbf{\#Frame} & \textbf{Video Embedder} & \textbf{Sound} & \textbf{Speech} & \textbf{Adapter} & \textbf{Hardware} & \textbf{LLM} & \textbf{LLM Size} & \textbf{Date} \\ \hline
VideoTree~\bluecite{wang2024videotree} & Varying & EVA-CLIP ViT-G/14 & \xmark & \xmark & - & 4 A6000 GPUs & GPT-4 & - & 04/2024 \\
\rowcolor[HTML]{ECF4FF} 
VTG-LLM~\bluecite{guo2024vtgllm} & 96 & EVA-CLIP ViT-G/14 & \xmark & \xmark & Projector & 6 ATN 910B & LLaMA-2 & 7B & 04/2024 \\
AutoAD III~\bluecite{han2024autoadiii} & 8 & EVA-CLIP & \cmark & \cmark & Linear Projector & 1 A40 GPU & GPT-3.5-turbo & - & 04/2024 \\
\rowcolor[HTML]{ECF4FF} 
LLaVA-Hound-DPO~\bluecite{zhang2024llavahounddpo} & 10 & LanguageBind ViT/14 & \xmark & \xmark & - & - & GPT-4V & - & 04/2024 \\
RedViLLM~\bluecite{huang2024redvillm} & Varying & CLIP ViT-G/14 & \xmark & \xmark & Temporal Module & - & Qwen-VL & 7B & 04/2024 \\
\rowcolor[HTML]{ECF4FF} 
LAVAD~\bluecite{zanella2024lavad} & 16 & BLIP-2 ViT-L/14 & \xmark & \xmark & - & - & LLaMA-2-chat & 13B & 04/2024 \\
VLM4HOI~\bluecite{bansal2024hoiref} & - & EVA CLIP & \xmark & \xmark & Projection Layer & 8 V100 GPUs & LLaMA-2-Chat & 13B & 04/2024 \\
\rowcolor[HTML]{ECF4FF} 
Koala~\bluecite{tan2024koala} & 64 & EVA-CLIP ViT-G/14 & \xmark & \xmark & Video QFormer+Linear Layer & 4 RTX A6000 GPUs & Vicuna-v0 & 7B/13B & 04/2024 \\
LongVLM~\bluecite{weng2024longvlm} & 100 & CLIP ViT-L/14 & \xmark & \xmark & Projection Layer & 4 A100 GPUs & Vicuna-v1.1 & 7B & 04/2024 \\
\rowcolor[HTML]{ECF4FF} 
MA-LMM~\bluecite{he2024malmm} & 100 & EVA-CLIP ViT-G/14 & \xmark & \xmark & Q-Former & 4 A100 GPUs & Vicuna & 7B & 04/2024 \\
MiniGPT4-Video~\bluecite{ataallah2024minigpt4video} & Varying & EVA-CLIP & \xmark & \xmark & Projector + LoRA for LLM & - & LlaMA-2/Mistral & 7B & 04/2024 \\
\rowcolor[HTML]{ECF4FF} 
MoReVQA~\bluecite{min2024morevqa} & 1 FPS & - & \xmark & \xmark & - & - & PaLM-2 & - & 04/2024 \\
Pegasus-v1~\bluecite{jung2024pegasusv1} & - & Marengo 2.6 & \xmark & \cmark & VL Alignment Module & - & GPT-4 & - & 04/2024 \\
\rowcolor[HTML]{ECF4FF} 
PLLaVA~\bluecite{xu2024pllava} & 16 & CLIP ViT-L/14 & \xmark & \xmark & MM Projector & - & LLaVA & 7B/13B/34B & 04/2024 \\
ST-LLM~\bluecite{liu2024stllm} & Varying & CLIP ViT-L/14 & \xmark & \xmark & Projection Layer & 8 A100 GPUs & Vicuna-v1.1 & 7B & 04/2024 \\
\rowcolor[HTML]{ECF4FF} 
Tarsier~\bluecite{wang2024tarsier} & Varying & CLIP ViT & \xmark & \xmark & MLP & 48 A100 GPUs & Vicuna & 7B/13B & 04/2024 \\
TraveLER~\bluecite{shang2024traveler} & Varying & - & \xmark & \xmark & - & 8 RTX A6000 GPUs & GPT-3.5 & - & 04/2024 \\
\rowcolor[HTML]{ECF4FF} 
V2Xum-LLaMA~\bluecite{hua2024v2xum} & 1 FPS & CLIP ViT-L/14 & \xmark & \xmark & Vision Adapter & 8 A100 GPUs & LLaMA & 7B/13B & 04/2024 \\
VITRON~\bluecite{fei2024vitron} & - & - & \xmark & \xmark & Projection Layer & 10 A100 GPUs & Vicuna-v1.5 & 7B & 04/2024 \\
\rowcolor[HTML]{ECF4FF} 
X-VARS~\bluecite{held2024xvars} & 16 & CLIP ViT-L/14 & \xmark & \xmark & Linear Layer & 1 V100 GPU & VideoChatGPT & - & 04/2024 \\
VideoChat2~\bluecite{li2024mvbench} & Varying & UMT-L & \xmark & \xmark & QFormer & - & Vicuna & 7B & 05/2024 \\
\rowcolor[HTML]{ECF4FF} 
Shotluck Holmes~\bluecite{luo2024shotluckholmes} & 120 & SigLip & \xmark & \xmark & MLP Projection Layer & 8 H100 GPUs & Vicuna & 7B & 05/2024 \\
VideoStreaming~\bluecite{qian2024videostreaming} & 16 & CLIP ViT-L/14 & \xmark & \xmark & Projecter & 32 A100 GPUs & Vicuna & 7B & 05/2024 \\
\rowcolor[HTML]{ECF4FF} 
VideoNarrator~\bluecite{yang2024videonarrator} & - & CLIP ViT-L/14 & \xmark & \xmark & Video Projecter & 4 A6000 GPUs & Baichuan & 7B & 05/2024 \\
TOPA~\bluecite{li2024topa} & 10 & CLIP ViT-L/14 & \xmark & \xmark & Linear Projector & 4 A100 GPUs & Llama2 & 7/8/13B & 05/2024 \\
\rowcolor[HTML]{ECF4FF} 
MotionLLM~\bluecite{chen2024motionllm} & 8 & LanguageBind/VQ-VAE & \xmark & \xmark & Linear Layer & A100 GPUs & Vicuna & 7B & 05/2024 \\
Artemis~\bluecite{qiu2024artemis} & Varying & CLIP ViT-L/14 & \xmark & \xmark & MLP & 8 A800 GPUs & Vicuna & 7B & 06/2024 \\
\rowcolor[HTML]{ECF4FF} 
DrVideo~\bluecite{ma2024drvideo} & 2 FPS & CLIP ViT-L/14 & \xmark & \xmark & - & - & LaViLa & 7B & 06/2024 \\
EmoLLM~\bluecite{yang2024emollm} & 8 & CLIP ViT-L/14 & \cmark & \cmark & Multi-Perspective Visual Projection & 4 4090 GPUs & Vicuna-v1.5 & 7B & 06/2024 \\
\rowcolor[HTML]{ECF4FF} 
FTFV-LLM~\bluecite{chen2024ftfvllm} & 8 & OpenCLIP ViT-G/14 & \xmark & \xmark & Vision-Language Adapter & 64 A100 GPUs & Vicuna & 7B & 06/2024 \\
Flash-VStream~\bluecite{zhang2024flashvstream} & 8 & - & \xmark & \xmark & MLP & 8 A100 GPUs & GPT-3.5 & - & 06/2024 \\
\rowcolor[HTML]{ECF4FF} 
Holmes-VAD~\bluecite{zhang2024holmesvad} & Varying & CLIP ViT-L/14 & \xmark & \xmark & Temporal Sampler & 2 A100 GPUs & Vicuna-v1.5 & 7B & 06/2024 \\
LongVA~\bluecite{zhang2024longva} & 1 FPS & CLIP ViT-L/14 & \xmark & \xmark & MLP & 8 A100 GPUs & Mistral-7B-Instruct-v0.2 & 7B & 06/2024 \\
\rowcolor[HTML]{ECF4FF} 
OmAgent~\bluecite{zhang2024omagent} & Varying & - & \xmark & \cmark & - & - & GPT-4o & - & 06/2024 \\
GIT-LLaVA~\bluecite{kalarani2024gitllava} & 6 & CLIP ViT-L/14 & \xmark & \xmark & MLP & 4 A100 GPUs & Vicuna & 7B & 06/2024 \\
\rowcolor[HTML]{ECF4FF} 
ShareGPT4Video~\bluecite{chen2024sharegpt4video} & 16 & CLIP ViT-L/14 & \xmark & \xmark & Linear Layer + LoRA for LLM & 8 A100 GPUs & LLaMA-3 & 8B & 06/2024 \\
LVNet~\bluecite{park2024lvnet} & - & - & \xmark & \xmark & - & - & GPT-4 & - & 06/2024 \\
\rowcolor[HTML]{ECF4FF} 
VIM~\bluecite{du2024vim} & - & EVA-CLIP ViT-G/14 & \xmark & \xmark & Q-former+LoRA in LLM & 8 A100 GPUs & Vicuna-v1.1 & 13B & 06/2024 \\
Video-SALMONN~\bluecite{sun2024videosalmonn} & 2 FPS & BLIP-2 ViT-G/14 & \cmark & \cmark & MRC Q-Former & - & GPT-4 & - & 06/2024 \\
\rowcolor[HTML]{ECF4FF} 
VideoGPT+~\bluecite{maaz2024videogpt+} & 16/8 & CLIP ViT-L/14 & \xmark & \xmark & Linear Layer & 8 A100 GPUs & Phi-3-Mini & 3.8B & 06/2024 \\
Vriptor~\bluecite{yang2024vript} & - & EVA-CLIP ViT-G/14 & \xmark & \cmark & - & A100 GPUs & - & - & 06/2024 \\
\rowcolor[HTML]{ECF4FF} 
MM-Screenplayer~\bluecite{wu2024mmscreenplayer} & - & - & \xmark & \cmark & - & - & GPT-4-turbo & - & 06/2024 \\
VideoLLaMA 2~\bluecite{cheng2024videollama2} & 16 & CLIP ViT-L/14 & \cmark & \xmark & Linear Layer +STC connector & - & LLaVA-1.5 & 7B & 06/2024 \\
\rowcolor[HTML]{ECF4FF} 
VideoLLM-online~\bluecite{chen2024videollmonline} & 2 FPS & CLIP ViT-L/14 & \xmark & \xmark & MLP projector + LoRA for LLM & - & Llama-2-Chat/Llama-3-Instruct & 7B/8B & 06/2024 \\ \hline
\end{tabular}%
}
\vspace{-1em}
\end{table*}
\subsubsection{Fine-tuning Vid-LLMs}
In contrast to most Vid-LLMs in the \textit{Video Analyzer $\times$ LLM} category being training-free, almost all Vid-LLMs in the \textit{Video Embedder $\times$ LLM} category undergo fine-tuning. The common methods for fine-tuning Vid-LLMs are categorized based on the types of adapters used during fine-tuning into four main types: LLM Fully Fine-tuning, Connective Adapter Fine-tuning, Insertive Adapter Fine-tuning, and Fine-tuning with Hybrid Adapters. An adapter is a small, trainable module added to a large model for fine-tuning. By only updating the parameters of these modules, the model can adapt to specific tasks without changing the entire model's parameters, achieving efficient parameter updates and task adaptation while conserving computational resources. Illustrations of each type are shown in \Cref{fig:finetuning}.
\begin{itemize}
    
    \item \textit{LLM Fully Fine-tuning:} This fine-tuning method does not use any adapters but instead employs supervised training with a lower learning rate, updating all the parameters in the LLM. This method allows the Vid-LLM to fully adapt to the respective task and achieve good performance, especially when the target task is quite different from the pretraining tasks. For end-to-end Vid-LLMs, especially those in the \textit{Video Embedder $\times$ LLM} category, the Video Embedder may also be fine-tuned for more comprehensive learning. However, this method consumes more computational resources than adapter-based fine-tuning methods and may potentially impair the inherent capabilities of the LLM, such as zero-shot and in-context learning. Vid-LLM adopted LLM Fully Fine-tuning include AV-LLM \bluecite{Shu2023AV-LLM} and Vid2Seq \bluecite{yang2023vid2seq}. In \bluecite{Shu2023AV-LLM}, there are both fully fine-tuning and adapter fine-tuning versions of Vid-LLMs, and the former's performance is better than the latter.

    \item \textit{Connective Adapter Fine-tuning:} Here, the term ``Connective" refers to adapters that bridge the Video Embedder and the LLM externally, enabling information from the video to flow into the LLM through the Connective Adapter. As illustrated in \Cref{fig:finetuning}, during training, the parameters of both the Video Embedder and the LLM are frozen, and only the parameters of the Connective Adapter are updated. Common Connective Adapters include MLP/Linear Layer and Q-former~\bluecite{li2023blip}, their combinations, etc., whose primary function is to map video embeddings from the visual semantic space to the text semantic space of the LLM input tokens (i.e., modality alignment). Typically, fine-tuning only the Connective Adapter does not alter the LLM's inherent behavior.
    

    \item \textit{Insertive Adapter Fine-tuning:} As the name suggests, Insertive Adapters are inserted into the LLM itself. Similar to using Connective Adapters, during training, the parameters of the Video Embedder and the LLM are frozen, and only the parameters of the Insertive Adapter are updated. Insertive Adapters, often based on LoRA, affect the LLM's behavior because they are added to the existing LLM parameters. This type of adapter is almost always present in Vid-LLMs classified as \textit{Video Embedder $\times$ LLM as Regressor} and \textit{Video Embedder $\times$ LLM as Hidden Layer}, as these types of Vid-LLMs require changes in the LLM's behavior, such as outputting continuous prediction values.

    \item \textit{Fine-tuning with Hybrid Adapters:} Many Vid-LLMs use a combination of Connective and Insertive Adapters to achieve both modality alignment and changes in the LLM's inherent behavior. Vid-LLMs employing Hybrid Adapters typically use multi-stage fine-tuning. A common approach is to fine-tune only the Connective Adapter in the first stage for modality alignment. In the second stage, the already fine-tuned Connective Adapter is frozen, the training task (from alignment task to target task) and the training data (from data used for modality alignment to data required for the target task) are changed, and only the parameters of the Insertive Adapter are updated. There are also single-stage approaches where both Connective and Insertive Adapters are updated simultaneously.
\end{itemize}

\section{Benchmarks and Evaluation}
\label{sec:benchmark_eval}
This section provides an overview of the evaluation methods for video question-answering models and related tasks, categorized into three types: closed-ended evaluation, open-ended evaluation, and other evaluation methods, which are shown in \Cref{tab:benchmarks}.
Closed-ended evaluations rely on questions with predefined answers or formats, including multiple-choice questions and structured formats that allow for straightforward scoring. Open-ended evaluations involve questions without predefined answer options, often requiring more sophisticated scoring methods, including LLM-based evaluations. Other evaluation methods address specialized video understanding capabilities such as temporal/spatiotemporal reasoning.

\subsection{Closed-ended Evaluation}
Closed-ended evaluations use pre-defined answers or structured formats~\bluecite{yin2023survey}. These include multiple-choice questions (TVQA~\bluecite{lei2018tvqa}, How2QA~\bluecite{How2QA}, STAR~\bluecite{wu2021star}) and questions with structured formats for direct comparison with ground truth (MSRVTT-QA~\bluecite{xu2017video}, MSVD-QA~\bluecite{xu2017video}, MVBench~\bluecite{li2024mvbench}). Multiple-choice performance is evaluated via accuracy percentages, while structured formats use metrics like CIDEr~\bluecite{vedantam2015cider}, METEOR~\bluecite{banerjee2005meteor}, ROUGE~\bluecite{lin2004rouge}, and SPICE~\bluecite{anderson2016spice} to compare predictions with ground truth.
Notable benchmarks include MSRVTT-QA~\bluecite{xu2017video}, TVQA~\bluecite{lei2018tvqa}, MVBench~\bluecite{li2024mvbench}, EgoSchema~\bluecite{mangalam2023egoschema}, and Video-MME~\bluecite{fu2024video-mme}. Each targets different video understanding aspects: TGIF-QA~\bluecite{jang2017tgif} focuses on action recognition, state transition, frame-level QA, and counting; ActivityNet-QA~\bluecite{yu2019activitynet} covers motion, spatial, temporal, and free-form dimensions; VidComposition~\bluecite{tang2024vidcomposition} emphasizes compositional reasoning; while NExT-QA~\bluecite{NExT-QA} and MLVU~\bluecite{zhou2024mlvu} include causal and temporal action reasoning.
These diverse question types test various reasoning abilities, though many benchmarks still exhibit domain biases toward common scenarios and lack diversity in rare events or unusual contexts.

\begin{table*}[]
\centering
\caption{The comparison of various benchmarks includes several important aspects: the total number of videos, the number of clips, the average duration of the videos, the number of QA pairs, and video content.}
\label{tab:benchmarks}
\resizebox{\linewidth}{!}{%
\begin{tabular}{l|lllllll}
\hline
\textbf{Benchmark} & \textbf{\#Videos} & \textbf{\#Clips} & \textbf{Len.(s)} & \textbf{Video Content} & \textbf{\#QA Pairs} & \textbf{Question Type} \\ \hline
MSRVTT-QA~\bluecite{xu2017video} & 2,990 & 2,990 & 15.2 & Open-domain & 72,821 & Closed-ended \& open-ended (what/who/how/when/where) \\
MSVD-QA~\bluecite{xu2017video} & 504 & 504 & 9.8 & Open-domain & 50,505 & Closed-ended \& open-ended (what/who/how/when/where) \\
TGIF-QA~\bluecite{jang2017tgif} & 9,575 & 9,575 & 3.0 & Open-domain & 8,506 & Closed-ended \& open-ended, action, transition, counting \\
ActivityNet-QA \bluecite{yu2019activitynet} & 800 & 800 & 111.4 & Human activity & 8,000 & Closed-ended \& open-ended (what/who/how/when/where/why) \\
TVQA~\bluecite{lei2018tvqa} & 2,179 & 15,253 & 11.2 & TV show & 15,253 & Closed-ended, multiple-choice \\
How2QA~\bluecite{How2QA} & 1,166 & 2,852 & 15.3 & TV episode & 2,852 & Multiple-choice \\
STAR~\bluecite{wu2021star} & 914 & 7,098 & 11.9 & Human action & 7,098 & Multiple-choice \\
NExT-QA~\bluecite{NExT-QA} & 1,000 & 1,000 & 39.5 & Daily life & 8,564 & Multiple-choice and open-ended (causal, temporal, descriptive) \\
MVBench~\bluecite{li2024mvbench} & 3,641 & 3,641 & 16.0 & Open-domain & 4,000 & Closed-ended (various tasks) \\
Video-Bench~\bluecite{ning2023video} & 5,917 & 5,917 & 56.0 & Open-domain & 17,036 & Open-ended (various tasks) \\
EgoSchema~\bluecite{mangalam2023egoschema} & 5,063 & 5,063 & 180.0 & Egocentric activity & 5,063 & Closed-ended, procedural understanding \\
AutoEval-Video~\bluecite{chen2023autoeval} & 327 & 327 & 14.6 & Open-domain & 327 & Open-ended evaluation \\
TempCompass~\bluecite{liu2024tempcompass} & 410 & 500 & 11.4 & Open-domain & 7,540 & Both closed-ended \& open-ended, temporal reasoning \\
Video-MME~\bluecite{fu2024video-mme} & 900 & 900 & 1,017.9 & Open-domain & 2,700 & Closed-ended evaluation \\
VideoVista~\bluecite{li2024videovista} & 894 & 3,400 & 131.0 & Open-domain & 25,000 & Open-ended (descriptive, causal, predictive) \\
CinePile~\bluecite{rawal2024cinepile} & 9,396 & 9,396 & 160.0 & Movie & 303,828 & Movie understanding, open-ended \\
SOK-Bench~\bluecite{wang2024sok} & 10,000 & 10,000 & - & Open-domain & 44,000 & Open-ended, subject-oriented knowledge \\
SFD~\bluecite{ghermi2024short} & 1,078 & 1,078 & 780.0 & Movies & 4,885 & Multiple-choice, open-ended \\
EditVid-QA~\bluecite{xu2024beyond} & 32,000 & 32,000 & - & Entertainment & 204,000 & Open-ended, editing techniques \\
InfiniBench~\bluecite{ataallah2024infinibench} & 1,219 & 1,219 & 4,460.4 & Movie/TV show & 108,200 & Closed-ended \& open-ended, long-form understanding \\
MLVU~\bluecite{zhou2024mlvu} & 1,334 & 1,334 & 180.0-7,200.0 & Open-domain & 2,593 & Both closed-ended and open-ended, long-form understanding \\
MMWorld~\bluecite{he2024mmworld} & 1,910 & 1,910 & 102.3 & Open-domain & 1,599 & Open-ended, multimodal evaluation \\
VELOCITI~\bluecite{saravanan2024velociti} & 900 & 900 & 10.0 & Movie & - & Open-ended, visual understanding \\
VidComposition~\bluecite{tang2024vidcomposition} & 982 & 982 & - & Movie/TV show & 1,706 & Multiple-choice, compositional reasoning \\
EAGLE~\bluecite{bi2024eagle}&36,000&36,000&16.0-76.0&Egocentric activity&400,000&Open-ended, procedural understanding\\
\hline
\end{tabular}%
}
\vspace{-1em}
\end{table*}

\subsection{Open-ended Evaluation}
Open-ended evaluation involves questions without pre-defined options or structured formats. While ground-truth answers serve as references, scoring methods are more sophisticated than option selection or string matching. GPT-3.5/4 models often evaluate predictions by comparing them with reference answers.
Notable open-ended benchmarks include MovieChat-1K~\bluecite{song2023moviechat}, MLVU~\bluecite{zhou2024mlvu}, NExT-QA~\bluecite{NExT-QA},  VELOCITI~\bluecite{saravanan2024velociti}, and EAGLE~\bluecite{bi2024eagle}. These require more complex responses demonstrating deeper reasoning. CinePile~\bluecite{rawal2024cinepile} incorporates analytical tasks like character dynamics and narrative analysis.
The most popular GPT-based evaluation methods, proposed in~\bluecite{maaz2023video}, are Open-end Zero-shot Video QA Evaluation and Video-based Generative Performance Benchmarking. Performance comparisons of Vid-LLMs on these metrics are shown in \Cref{tab:zero-shot}.
Originally closed-ended benchmarks like MSRVTT-QA~\bluecite{xu2017video}, MSVD-QA~\bluecite{xu2017video}, TGIF-QA~\bluecite{jang2017tgif}, and ActivityNet-QA~\bluecite{yu2019activitynet} can be repurposed as open-ended in GPT-based evaluations, as LLMs generate free-form responses that GPT models compare to references.
These methods have limitations: evaluation scores change with GPT version updates, making cross-study comparisons difficult; results depend heavily on prompt engineering; and LLM evaluators may favor responses similar to their generation patterns rather than objectively assessing quality.

\subsection{Other Evaluations}
Other benchmarks evaluate fine-grained temporal and spatiotemporal understanding. Dense captioning generates~\bluecite{shao2022region,Wang_2021_ICCV,shao2023textual,long2023capdet,shao2024dcmstrd} detailed descriptions for multiple video events/objects, using BLEU, METEOR, and CIDEr metrics that assess both temporal localization and descriptive accuracy. Vid-LLMs' performance of dense video captioning on ActivityNet Captions~\bluecite{krishna2017dense} is shown in \Cref{tab:dvc_comparison}.
Several Vid-LLMs have already achieved performance comparable to traditional task-specific models in dense video captioning.
Video temporal grounding localizes specific moments based on textual queries, evaluated using tIoU and Recall@K. Spatiotemporal grounding extends this to localize in both space and time, assessed via spatiotemporal IoU and mAP.
Object tracking~\bluecite{wang2024elysium, Li2024GroundingGPT} follows objects across frames, evaluated using precision, success rate, and tracking accuracy. Video saliency detection~\bluecite{tang2024cardiff} identifies visually salient regions, evaluated with AUC-J, NSS, etc.
These tasks rely on temporal or spatiotemporal annotations as ground-truth, with metrics like IoU, Recall@K, and mAP widely adopted. Human evaluation is also used for subjective aspects, though this is labor-intensive and time-consuming.

As for qualitative evaluation, several approaches can effectively assess Vid-LLMs' performance in addition to numerical metrics. 
Error analysis~\bluecite{li2024mvbench,hua2024mmcomposition} for open-ended QA and difference comparisons~\bluecite{huang2023vtimellm, tang2024avicuna}  between model outputs and ground truth annotations (\textit{e.g.}, intervals) for temporal/spatiotemporal understanding provide insights into model limitations. Attention visualization~\bluecite{bi2024unveiling} reveals what visual elements the models prioritize when generating responses. Self-explanation~\bluecite{hua2024mmcomposition,tang2024vidcomposition}, where models justify their answers for closed-ended benchmarks, offers valuable insights into reasoning processes and potential misconceptions. Human studies, though resource-intensive, remain helpful in finding models that reflect human preferences.

\begin{table}[h]
    \centering
        \caption{Comparison of Vid-LLMs and conventional models (non-LLM-based) on dense video captioning models on ActivityNet Captions dataset.}
    \label{tab:dvc_comparison}
    \resizebox{0.95\columnwidth}{!}{%
    \begin{tabular}{lcccc}
        \hline
        \textbf{Model} & \textbf{CIDEr} & \textbf{SODA$_c$} & \textbf{METEOR} & \textbf{F1} \\
        \hline
        \textit{Non-LLM-based} & & & & \\
        MT~\bluecite{zhou2018end-to-end} & 9.3 & -- & 5.0 & -- \\
        PDVC~\bluecite{Wang_2021_ICCV} & 29.3 & 6.0 & 7.6 & -- \\
        CM$^2$~\bluecite{kim2024cm2} & 33.1 & 6.2 & 8.5 & 54.2 \\
        \hline
        \textit{LLM-based} & & & & \\
        Momentor~\bluecite{Qian2024Momentor} & 14.9 & 2.3 & 4.7 & -- \\
        TimeChat~\bluecite{Ren2024TimeChat} & 19.0 & 4.7 & 5.7 & 36.9 \\
        VTG-LLM~\bluecite{guo2024vtgllm} & 20.7 & 5.1 & 5.9 & 34.8 \\
        AVicuna~\bluecite{tang2024avicuna} & 22.5 & 5.1 & 6.5 & 45.2 \\
        TRACE~\bluecite{guo2024trace} & {25.9} & 6.0 & 6.4 & 39.3 \\
        VTimeLLM~\bluecite{huang2024vtimellm} & 27.6 & 5.8 & 6.8 & -- \\
        TRACE-uni~\bluecite{guo2024trace} & 29.2 & 6.4 & 6.9 & 40.4 \\
        GIT~\bluecite{wang2022git} & 29.8 & 5.7 & 7.8 & 50.6 \\
        Vid2Seq~\bluecite{yang2023vid2seq} & 30.2 & 5.9 & 8.5 & 51.8 \\
        Streaming Vid2Seq~\bluecite{zhou2024streamingdvc} & 37.8 & 6.2 & 10.0 & {52.9} \\
        Streaming GIT~\bluecite{zhou2024streamingdvc} & 41.2 & 6.6 & 9.0 & 50.9 \\
        \hline
    \end{tabular}%
    }
\vspace{-1em}
\end{table}

\begin{table*}[]
\centering
\caption{This table comprehensively compares various Vid-LLMs across multiple open-end zero-shot video question answering and video-based generative performance benchmarks. It includes GPT-based metrics for MSVD-QA, MSRVTT-QA, and ActivityNet-QA datasets, as well as scores for Correctness of Information, Detail Orientation, Contextual Understanding, Temporal Understanding, and Consistency aspects in video-based generative performance.}
\label{tab:zero-shot}
\resizebox{0.95\textwidth}{!}{%
\begin{tabular}{l|ccc|cccccc}
\hline
 & \multicolumn{3}{c|}{\textbf{Open-end Zero-shot Video QA Evaluation}} & \multicolumn{6}{c}{\textbf{Video-based Generative Performance Benchmarking}} \\ \cline{2-10} 
\multirow{-2}{*}{\textbf{Model}} & MSVD-QA & MSRVTT-QA & ActivityNet-QA & Correctness & Detail & Context & Temporal & Consistency & Average \\ \hline
GPT4-V~\bluecite{gpt4v2023online} & - & - & 59.5 & 4.09 & 3.88 & 4.37 & 3.94 & 4.02 & 4.06 \\
Video-LLaMA~\bluecite{zhang2023video} & 51.6 & 29.6 & 12.4 & 1.96 & 2.18 & 2.16 & 1.82 & 1.79 & 1.98 \\
LLaMA-Adapter~\bluecite{gao2023llama} & 54.9 & 43.8 & 34.2 & 2.03 & 2.32 & 2.30 & 1.98 & 2.15 & 2.16 \\
VideoChat~\bluecite{li2023videochat} & 56.3 & 45.0 & 26.5 & 2.23 & 2.50 & 2.53 & 1.94 & 2.24 & 2.29 \\
VALLY~\bluecite{luo2023valley} & 60.5 & 51.1 & 45.1 & 2.43 & 2.13 & 2.86 & 2.04 & 2.45 & 2.38 \\
MovieLLM~\bluecite{chen2023videollm} & 63.2 & 52.1 & 43.3 & 2.64 & 2.61 & 2.92 & 2.03 & 2.43 & 2.53 \\
Video-ChatGPT~\bluecite{maaz2023video} & 64.9 & 49.3 & 35.2 & 2.50 & 2.57 & 2.69 & 2.16 & 2.20 & 2.42 \\
Chat-UniVi~\bluecite{Jin2024Chat-UniVi} & 65.0 & 54.6 & 46.1 & 2.89 & 2.91 & 3.46 & 2.89 & 2.81 & 2.99 \\
Vista-LLaMA~\bluecite{Ma2023Vista-LLaMA} & 65.3 & 60.5 & 48.3 & 2.44 & 2.64 & 3.18 & 2.26 & 2.31 & 2.57 \\
AV-LLM~\bluecite{Shu2023AV-LLM} & 67.3 & 53.7 & 47.2 & 2.56 & 2.47 & 2.93 & 2.17 & 2.47 & 2.52 \\
LLaVA-NeXT-Video~\bluecite{zhang2024llavanextvideo} & 67.8 & - & 53.5 & 3.39 & 3.29 & 3.92 & 2.60 & 3.12 & 3.26 \\
LLaMA-VID~\bluecite{li2023llama-vid} & 69.7 & 57.7 & 47.4 & 2.96 & 3.00 & 3.53 & 2.46 & 2.51 & 2.89 \\
VTimeLLM~\bluecite{huang2023vtimellm} & 69.8 & 58.8 & 45.5 & 2.78 & 3.10 & 3.40 & 2.49 & 2.47 & 2.85 \\
VideoChat2~\bluecite{li2024mvbench} & 70.0 & 54.1 & 49.1 & 3.02 & 2.88 & 3.51 & 2.66 & 2.81 & 2.98 \\
LongVLM~\bluecite{weng2024longvlm} & 70.0 & 59.8 & 47.6 & 2.76 & 2.86 & 3.34 & 2.39 & 3.11 & 2.89 \\
AVicuna~\bluecite{tang2024avicuna} & 70.2 & 59.7 & 53.0 & 2.81 & 2.62 & 3.25 & 2.53 & 2.59 & 2.76 \\
Video-LLaVA~\bluecite{lin2023video} & 70.7 & 59.2 & 45.3 & 2.87 & 2.94 & 3.44 & 2.45 & 2.51 & 2.84 \\
RED-VILLM~\bluecite{huang2024redvillm} & 71.2 & 53.9 & 44.2 & 2.71 & 2.75 & 3.34 & 2.34 & 2.45 & 2.72 \\
Video LLaMA 2~\bluecite{cheng2024videollama2} & 71.7 & - & 49.9 & 3.09 & 3.09 & 3.68 & 2.63 & 3.25 & 3.15 \\
Artemis~\bluecite{qiu2024artemis} & 72.1 & 56.7 & 39.3 & 2.69 & 2.55 & 3.04 & 2.24 & 2.70 & 2.64 \\
VideoGPT+~\bluecite{maaz2024videogpt+} & 72.4 & 60.6 & 50.6 & 2.46 & 2.73 & 2.81 & 1.78 & 2.47 & 2.45 \\
MiniGPT4-video~\bluecite{ataallah2024minigpt4video} & 73.9 & 58.8 & 44.4 & 3.09 & 3.02 & 3.57 & 2.56 & 2.67 & 2.98 \\
ST-LLM~\bluecite{liu2024stllm} & 74.6 & 63.2 & 50.9 & 3.23 & 3.05 & 3.74 & 2.93 & 2.81 & 3.15 \\
MovieChat~\bluecite{song2023moviechat} & 75.2 & 52.7 & 45.7 & 2.76 & 2.93 & 3.01 & 2.24 & 2.42 & 2.67 \\
PLLaVA~\bluecite{xu2024pllava} & 76.6 & 62.0 & 56.3 & 3.21 & 2.86 & 3.62 & 2.33 & 2.93 & 2.99 \\
IG-VLM~\bluecite{Kim2024IG-VLM} & 76.7 & 62.7 & 57.3 & 3.26 & 2.76 & 3.57 & 2.34 & 3.28 & 3.04 \\ \hline
\end{tabular}%
}
\vspace{-1em}
\end{table*}

\subsection{Analysis of Model Performance} \label{sec:model_attribute_analysis}
Analyzing the correlation between model attributes and benchmark performance reveals several key factors driving recent improvements in Vid-LLMs. From Tables \ref{tab:dvc_comparison} and \ref{tab:zero-shot}, we observe that models built on larger and more recent foundation LLMs (e.g., IG-VLM with 34B parameters) consistently outperform their smaller counterparts, particularly in zero-shot VideoQA tasks. Models employing more powerful visual embedders such as EVA-CLIP or ViT-G architectures (notably in PLLaVA, IG-VLM, and Video LLaMA 2) demonstrate superior performance across both dense captioning and QA benchmarks. The frame sampling strategy also significantly impacts results, with high performers on temporal tasks (like VTimeLLM, AVicuna, and ST-LLM) typically processing more frames (100+) than general understanding models, while sophisticated adaptation mechanisms beyond simple projection layers (such as Q-formers or cross-attention) contribute to better contextual understanding. Performance gains stem from a combination of stronger foundation models, better visual encoders, appropriate temporal modeling, and more sophisticated bridging architectures rather than any single innovation.
\section{Applications and Future Directions}
\label{sec:sum}

\subsection{Application Scenarios}
Vid-LLMs have revolutionized various sectors by enabling advanced video and language processing capabilities. We outlines their diverse applications, demonstrating the extensive and transformative impact of Vid-LLMs across industries.
\subsubsection{Media and Entertainment}
\begin{itemize}
    \item \textit{Online Video Platforms and Multimedia Information Retrieval:} Vid-LLMs significantly enhance search algorithms~\bluecite{mao2023large}, generate context-aware video recommendations~\bluecite{ju2022prompting}, and aid in natural language tasks such as subtitle generation and translation~\bluecite{yang2023vid2seq}, contributing to online video platforms and multimedia retrieval systems. Their capabilities in analyzing videos for specific keyword retrieval~\bluecite{zhao2023lavila, jin2023text,jin2023diffusionret} improve intelligent recommendation systems. In the multimedia fields, it combines videos in domains like music~\bluecite{xu2023launchpadgpt}, avatar~\bluecite{song2021tacr,song2023emotional,song2024tri,song2021fsft}, and scene~\bluecite{Song_2023_CVPR}, to assist with content generation.

    \item \textit{Video Summarization and Editing:} Vid-LLMs are integral in generating concise summaries of video content~\bluecite{Pramanick_2023_ICCV}, which analyzes visual and auditory elements to extract key features for context-aware summaries. They also contribute to the field of video editing, as covered in existing literature~\bluecite{wu2023visual} and advertisement editing~\bluecite{tang2022multi}.
\end{itemize}

\subsubsection{Interactive and User-Centric Systems}
\begin{itemize}
    \item \textit{Virtual Education, Accessibility, and Sign Language:} Vid-LLMs serve as virtual tutors in education, analyzing instructional videos for interactive learning environments~\bluecite{gan2023large}. They also facilitate sign language translation into spoken language or text~\bluecite{liu2023survey, de2023machine}, improving accessibility for the deaf and hard of hearing.

    \item \textit{Interactive Gaming and Virtual Environments:} In the gaming industry, Vid-LLMs play a crucial role in creating dynamic dialogues and storylines, as well as aiding in generating procedural content, such as quests and in-game texts~\bluecite{mishra2023generating,koomen2023text}. They also power customer service chatbots~\bluecite{soni2023large,medeiros2023analysis}. Additionally, in AR/VR/XR, Vid-LLMs contribute to the generation of dynamic narrative content, enhancing user immersion~\bluecite{gokce2023role,jung2023xr,yu2024promptfix,huang2023egocentric}.

    \item \textit{State-Aware Human-Computer Interaction and Robot Planning:} In the field of human-computer interaction, Vid-LLMs analyze user videos to discern context and provide customized assistance, as highlighted in Bi et al.~\bluecite{bi2023misar}. Interaction forms also involve video content understanding like captioning videos~\bluecite{wang2023caption, hu2022promptcap, hua2024finematch}. Concurrently, in autonomous robot navigation, the SayPlan method~\bluecite{rana2023sayplan} integrates LLMs with 3D scene graphs to enable robots to interpret and navigate complex spaces in large buildings. 
\end{itemize}

\subsubsection{Healthcare and Security Applications}
\begin{itemize}
    \item \textit{Healthcare Innovations:} In the healthcare sector, Vid-LLMs play a crucial role in processing and interpreting medical literature, assisting in diagnostic and educational processes~\bluecite{eysenbach2023role,liu2022beat,liu2022disco,liu2024emage}, and providing decision support for healthcare professionals. They are used in patient interaction tools, such as chatbots for symptom assessment and addressing health-related queries, thus improving patient care and access to information~\bluecite{li2023llava}.
    
    \item \textit{Security, Surveillance, and Cybersecurity:} Vid-LLMs are crucial in security and protection, analyzing communications for potential threats~\bluecite{al2023chatgpt,mouratidis2023modelling} and detecting anomalous patterns in data~\bluecite{lee2023lanobert,almodovar2023logfit}. In surveillance video analysis, they identify suspicious behaviors, helping law enforcement~\bluecite{de2023socratic}. Their role in cybersecurity includes identifying phishing attempts and contributing to forensic analysis by summarizing case-related texts~\bluecite{tang2023graphgpt}. They may also improve video crowd counting~\bluecite{cao2025efficientmaskedautoencodervideo} for security applications.

    \item \textit{Autonomous Vehicles:} In autonomous vehicles, Vid-LLMs can process language inputs for interaction~\bluecite{cui2024drive}, assist in understanding road signs and instructions~\bluecite{li2023otter,lai2023lisa}, and improve user interfaces for vehicle control systems~\bluecite{cui2024drive}, enhancing safety and user experience.
\end{itemize}

\subsubsection{Other Applications}
Vid-LLMs offer valuable applications beyond those previously discussed. In video generation research~\bluecite{zhou2024survey,lin2023videodirectorgpt,kondratyuk2023videopoet}, Vid-LLMs can evaluate model performance, refine text prompts, and provide reasoning capabilities that better reflect human intentions. Additionally, Vid-LLMs show promise in resource-constrained environments through edge computing applications~\bluecite{jin2024efficient, lu2024b, hu2023edge} and can enhance privacy-preserving distributed systems through federated learning frameworks~\bluecite{yao2024federated, bastola2024fedmil, wang2023fedvmr}.

\subsection{Future Directions}
Despite enhancing multiple downstream tasks, Vid-LLMs face several challenges in real-world video understanding:
    \subsubsection{More Fine-grained Video Understanding} Fine-grained understanding remains challenging due to limited datasets, insufficient research, and high computational demands. The frame-by-frame analysis increases computational load while capturing spatiotemporal information. Understanding deeper semantics (emotions, scene dynamics) is harder, though text-video alignment through LLMs offers promise~\bluecite{tang2024vidcomposition}.  
    
    \subsubsection{Long-form Video Understanding} Long videos' extended duration complicates the analysis, especially in understanding events over time.  Thus, identifying key events and maintaining attention in long videos is difficult~\bluecite{zhang2024longva, weng2024longvlm, Wang2024LifelongMemory}. Effective mechanisms are needed to detect and highlight important parts, particularly in content-rich or complex plot videos.
    
    \subsubsection{Multimodal Video Understanding} Multimodal video understanding requires integrating different types of data, such as visual, audio, and text, for a better understanding of videos~\bluecite{tang2024avicuna,mohammadkhani2025survey,zhangmultimedia}. Aligning these data, especially in terms of spatial and temporal synchronization, is particularly critical. This area lacks relevant research and suffers from a scarcity of datasets. The field lacks research and datasets, with challenges in ensuring high-quality data annotation.

    \subsubsection{Hallucination in Video LLMs} Hallucination occurs when models generate responses disconnected from source material~\bluecite{zhang2024eventhallusion}, caused by insufficient feature extraction, influence of video context, domain gap between vision and language, and inherent LLM hallucinations. Solutions include post-training strategies~\bluecite{zhang2024llavahounddpo}, enhanced spatiotemporal context understanding, and visual-linguistic latent collaboration.

    \subsubsection{Industrial Deployment and Scalability} Effective deployment strategies~\bluecite{weng2024longvlm, lee2024video, tang2024enhancing, tan2024koala, shang2024interpolating, lu2024b} include model compression, token merging, domain-specific fine-tuning, modular architectures, efficient caching, and standardized integration frameworks, balancing efficiency with performance for industrial systems.

\subsection{Ethical Implications}
The ethical implications of Vid-LLMs center on privacy, data security, and potential misuse.
These models perform tasks like video engagement analysis, transcription, summarization, and captioning, requiring access to sensitive content.
This raises privacy risks, as video data may contain private or confidential information that could be exposed without proper consent.
Also, Vid-LLMs can be misused for surveillance or generating misleading content.
Bias is another concern, especially if training data lacks diversity.
Addressing these issues requires robust data governance, consent mechanisms, and ethical deployment to prioritize privacy and fairness.

\section{Conclusion}

This survey has examined the integration of LLMs in video understanding, which has enabled more sophisticated and versatile processing capabilities beyond traditional methods.
We categorized current approaches into three main types: \textit{Video Analyzer $\times$ LLM}, \textit{Video Embedder $\times$ LLM}, and \textit{(Analyzer + Embedder) $\times$ LLM}, with sub-classifications based on LLM functional roles: \textit{Summarizer}, \textit{Manager}, \textit{Text Decoder}, \textit{Regressor}, and \textit{Hidden Layer}.
Vid-LLMs demonstrate capabilities in multi-granularity reasoning from abstract to spatiotemporal analysis, showing potential across video summarization, captioning, question answering, and other applications. Despite progress, limitations remain in evaluation metrics, long-form video handling, and visual-textual modality alignment.
Future research will address these challenges through more efficient training strategies, improved Vid-LLM scalability, innovative architectures for multimodal integration, enhanced long-form video understanding, and methods to mitigate hallucinations. Expanding datasets and benchmarks will be critical for advancing video understanding with LLMs.


\bibliographystyle{IEEEtran}


\bibliography{reference}    

\newpage

\vspace{-33pt}
\begin{IEEEbiographynophoto}
{Yunlong Tang} received the B.Eng. degree in Intelligence Science and Technology from the Southern University of Science and Technology (SUSTech) in 2023, supervised by Prof. Feng Zheng. She is currently pursuing a Ph.D. degree in Computer Science at the University of Rochester, advised by Prof. Chenliang Xu. Her research focuses on multimodal learning, especially video understanding and generation. 
\end{IEEEbiographynophoto}
\vspace{-33pt}
\begin{IEEEbiographynophoto}
{Jing Bi} is currently pursuing a Ph.D. in Computer Science at the University of Rochester since 2020, advised by Prof. Chenliang Xu. He received his B.S. from Shandong University and M.S. from the University of Rochester.
\end{IEEEbiographynophoto}
\vspace{-33pt}
\begin{IEEEbiographynophoto}{Siting Xu} received her B.Eng. (2019 - 2023) degree in Computer Science and Technology from Southern University of Science and Technology (SUSTech), supervised by Prof. Feng Zheng.
\end{IEEEbiographynophoto}
\vspace{-33pt}
\begin{IEEEbiographynophoto}{Luchuan Song} is currently a Ph.D. candidate in Computer Science at the University of Rochester. He received his M.S. and B.S. from University of Science and Technology of China.
\end{IEEEbiographynophoto}
\vspace{-33pt}
\begin{IEEEbiographynophoto}{Susan Liang} is currently a Ph.D. candidate in the Computer Science Department at the University of Rochester. His research focuses on multimodal learning.
\end{IEEEbiographynophoto}
\vspace{-33pt}
\begin{IEEEbiographynophoto}{Teng Wang} is currently a Ph.D. candidate in the Department of Computer Science at the University of Hong Kong. He obtained his bachelor's and master's degrees from Sun Yat-sen University in 2017 and 2020, respectively. His research interests lie in vision-language multimodal learning and video understanding.
\end{IEEEbiographynophoto}
\vspace{-33pt}

\begin{IEEEbiographynophoto}{Daoan Zhang} is currently a Ph.D. Student in Computer Science at the University of Rochester, advised by Prof. Jiebo Luo. His research focuses on generative AI.
\end{IEEEbiographynophoto}
\vspace{-33pt}

\begin{IEEEbiographynophoto}{Jie An} is a Ph.D. candidate in Computer Science at the University of Rochester, advised by Prof. Jiebo Luo. Prior to that, he obtained his B.S. (2012 - 2016) and M.S. (2016 - 2019) in Applied Mathematics from Peking University, advised by Prof. Jinwen Ma. Jie's research focuses on improving the performance and extending the capability of GenAI models. He is particularly interested in image style transfer, generative models, image/video generation, and multi-modal generation/evaluation.
\end{IEEEbiographynophoto}
\vspace{-33pt}
\begin{IEEEbiographynophoto}{Jingyang Lin} is a PhD student majoring in Computer Science at the University of Rochester, advised by Professor Jiebo Luo. He received his BE and MSc degrees from Sun Yat-sen University (SYSU), Guangzhou, China, in 2019 and 2022, respectively. His research interests include multimodal learning with LLMs, AI for health, and self-supervised learning.
\end{IEEEbiographynophoto}
\vspace{-33pt}

\begin{IEEEbiographynophoto}{Rongyi Zhu} is currently a Ph.D. student in Computer Science at Stony Brook University. He received his MS. degree in Computer Science from the University of Rochester in 2024. His research focuses on trustworthy AI.
\end{IEEEbiographynophoto}
\vspace{-33pt}
\begin{IEEEbiographynophoto}{Ali Vosoughi} is a PhD candidate in Electrical and Computer Engineering at the University of Rochester, working with Professors Chenliang Xu and Axel Wismueller. His research focuses on using AI for multimodality and complex reasoning to assist humans with challenging tasks. He holds a Bachelor's degree in Electrical Engineering from Sharif University of Technology (Iran), and two Master's degrees—one from Bogazici University (Turkey) and another from the University of Rochester (USA).
\end{IEEEbiographynophoto}
\vspace{-33pt}
\begin{IEEEbiographynophoto}{Chao Huang} is currently a PhD candidate in Computer Science at the University of Rochester, advised by Prof. Chenliang Xu. Previously, he spent a year as a research assistant at the Chinese University of Hong Kong, working with Prof. Chi-Wing Fu. He received his B.Eng. from ESE Department, Nanjing University in 2019.
\end{IEEEbiographynophoto}
\vspace{-33pt}
\begin{IEEEbiographynophoto}{Zeliang Zhang} received a B.Eng. degree in computer science from Huazhong University of Science and Technology in 2022. He is currently a Ph.D. candidate in Computer Science at the University of Rochester. His research focuses on trustworthy and efficient deep learning methods.
\end{IEEEbiographynophoto}
\vspace{-33pt}
\begin{IEEEbiographynophoto}{Pinxin Liu} is a Ph.D. student at the University of Rochester, advised by Prof. Chenliang Xu. Before that, he was a Research Scientist Intern at Flawless AI. He received his B.S. from the Department of Computer Science at the University of Rochester. His research interest lies in human-related topics, e.g., video gesture synthesis, 3D face rendering, and text-to-motion generation.
\end{IEEEbiographynophoto}
\vspace{-33pt}
\begin{IEEEbiographynophoto}{Mingqian Feng} received his B.S. degree in Physics from University of Science and Technology of China (USTC) and his M.S.E. degree in financial mathematics from Johns Hopkins University. He is currently a Ph.D. student in Computer Science at the University of Rochester. His research focuses on bias, optimization, and video understanding.
\end{IEEEbiographynophoto}
\begin{IEEEbiographynophoto}{Feng Zheng} is an Associate Professor at Southern University of Science and Technology (SUSTech). His research interests include machine learning (ML), computer vision (CV) and human-computer interaction (HCI). He received a Ph.D. from the University of Sheffield, UK. Before joining SUSTech, he worked as a senior researcher at Tencent YouTu Lab in Shanghai, China. Prior to this, he worked as a postdoctoral researcher at the University of Pittsburgh, USA and as an assistant research professor at the Shenzhen Institute of Advanced Technology, CAS. In terms of academic research, he has published 85 papers in top international journals and conferences, including IEEE TPAMI/TITS/TIP, AAAI, NeuIPS, CVPR, and ECCV.
\end{IEEEbiographynophoto}
\vspace{-33pt}
\begin{IEEEbiographynophoto}{Jianguo Zhang} is a Professor in the Department of Computer Science and Engineering, Southern University of Science and Technology. Previously, he was a Reader in Computing, School of Science and Engineering, University of Dundee, UK. He received a PhD from the National Lab of Pattern Recognition, Institute of Automation, Chinese Academy of Sciences, Beijing, China, 2002. His research interests include object recognition, medical image analysis, machine learning, and computer vision. He is a senior member of the IEEE and an Associate Editor of IEEE Transactions on Multimedia.
\end{IEEEbiographynophoto}
\vspace{-33pt}
\begin{IEEEbiographynophoto}{Ping Luo} is an Associate Professor in the Department of Computer Science at the University of Hong Kong, Associate Director of the HKU Musketeers Foundation Institute of Data Science, and Deputy Director of the Joint Research Lab of HKU and Shanghai AI Lab. He earned his Ph.D. in Information Engineering from the Chinese University of Hong Kong in 2014, supervised by Prof. Xiaoou Tang and Prof. Xiaogang Wang. Before joining HKU in 2019, he was a Research Director at SenseTime. He has published over 100 papers in top conferences and journals, with 50,000+ citations on Google Scholar. His awards include the 2015 AAAI Easily Accessible Paper, 2022 ACL Outstanding Paper, 2023 WAIC Outstanding Paper, and ICCV’23 Best Paper nomination. In 2020, he was named one of the MIT Technology Review's Innovators Under 35 in Asia-Pacific. He has mentored 30 Ph.D. students, many of whom have won prestigious awards like the Nvidia and Baidu Fellowships.
\end{IEEEbiographynophoto}
\vspace{-33pt}
\begin{IEEEbiographynophoto}{Jiebo Luo} is the Albert Arendt Hopeman Professor of Engineering at the University of Rochester, which he joined in 2011 after over 15 years at Kodak Research Laboratories. He has played key roles in numerous conferences, serving as General Co-Chair of ACM Multimedia 2018 and IEEE ICME 2024, and Program Co-Chair of ACM Multimedia 2010, IEEE CVPR 2012, and IEEE ICIP 2017. He has served on editorial boards of several major journals, including IEEE Transactions on Pattern Analysis and Machine Intelligence and IEEE Transactions on Multimedia, and was the Editor-in-Chief of IEEE Transactions on Multimedia from 2020 to 2022. Prof. Luo is a Fellow of ACM, AAAI, IEEE, AIMBE, IAPR, and SPIE, and a member of Academia Europaea (AE) and the US National Academy of Inventors (NAI). His honors include the 2024 IEEE Computer Society Edward J. McClusky Technical Achievement Award, the 2021 ACM SIGMM Technical Achievement Award, the 2024 William H. Riker University Award for Excellence in Graduate Teaching, the 2024 Edmund A. Hajim Outstanding Faculty Award, and the inaugural 2024 Debra Haring Excellence in Research Award.
\end{IEEEbiographynophoto}
\vspace{-33pt}
\begin{IEEEbiographynophoto}{Chenliang Xu} is an Associate Professor in the Department of Computer Science at the University of Rochester. He received his Ph.D. in Computer Science from the University of Michigan in 2016, an M.S. in Computer Science from the University at Buffalo in 2012, and a B.S. in Information and Computing Science from Nanjing University of Aeronautics and Astronautics, China, in 2010. His research originates in computer vision and tackles interdisciplinary topics, including video understanding, audio-visual learning, vision and language, and methods for trustworthy AI. Xu is a recipient of the James P. Wilmot Distinguished Professorship (2021), the University of Rochester Research Award (2021), the Best Paper Award at the 17th ACM SIGGRAPH VRCAI Conference (2019), the Best Paper Award at the 14th Sound and Music Computing Conference (2017), and the University of Rochester AR/VR Pilot Award (2017). He has authored over 100 peer-reviewed papers in computer vision, machine learning, multimedia, and AI venues. He served as an associate editor for IEEE Transactions on Multimedia and area chair/reviewer for various international conferences.
\end{IEEEbiographynophoto}

\vfill

\end{document}